\documentclass[10pt,twocolumn,letterpaper]{article}

\usepackage[preprint]{cvpr}

\definecolor{cvprblue}{rgb}{0.21,0.49,0.74}
\usepackage[pagebackref,breaklinks,colorlinks,allcolors=cvprblue]{hyperref}

\def\paperID{17147} 
\def\confName{CVPR}
\def\confYear{2025}
\usepackage[utf8]{inputenc} 
\usepackage[T1]{fontenc}    
\usepackage{booktabs}       
\usepackage{amsfonts}       
\usepackage{nicefrac}       
\usepackage{microtype}      
\usepackage{amsmath}

\usepackage{amssymb}
\usepackage{pgfplots}
\usetikzlibrary{pgfplots.groupplots}
\usepackage{multirow}
\usepackage{bbding}
\usepackage{pifont}
\usepackage{caption}
\usepackage{svg}
\usepackage{wrapfig}
\usepackage{bm}

 \usepackage{microtype}
 \usepackage{booktabs}
\usepackage{algorithm}
 \usepackage{algpseudocode}

 \usepackage{colortbl} 
 \usepackage{xcolor}
 \usepackage{graphicx}

\usepackage{comment}

\definecolor{tabfirst}{rgb}{1, 0.7, 0.7} 
\definecolor{tabsecond}{rgb}{1, 0.85, 0.7} 
\definecolor{tabthird}{rgb}{1, 1, 0.7} 

\usepackage{hyperref}

\title{Toward Robust Neural Reconstruction from Sparse Point Sets}

\author{Amine Ouasfi, 
Shubhendu Jena, 
Eric Marchand, 
Adnane Boukhayma\\
Inria, Univ. Rennes, CNRS, IRISA}

\begin{document}

\maketitle

\begin{abstract}
We consider the challenging problem of learning Signed Distance Functions (SDF) from sparse and noisy 3D point clouds. 
In contrast to recent methods that depend on smoothness priors, our method, rooted in a distributionally robust optimization (DRO) framework, incorporates a regularization term that leverages samples from the uncertainty regions of the model to improve the learned SDFs. Thanks to tractable dual formulations, we show that this framework enables a stable and efficient optimization of SDFs in the absence of ground truth supervision. Using a  variety of synthetic and real data evaluations from different modalities, we show that our DRO based learning framework can improve SDF learning with respect to baselines and the state-of-the-art methods. 
\end{abstract}

\section{Introduction}
\label{sec:intro}

\begin{figure}[t]
\centering
\includegraphics[width=1.0\linewidth]{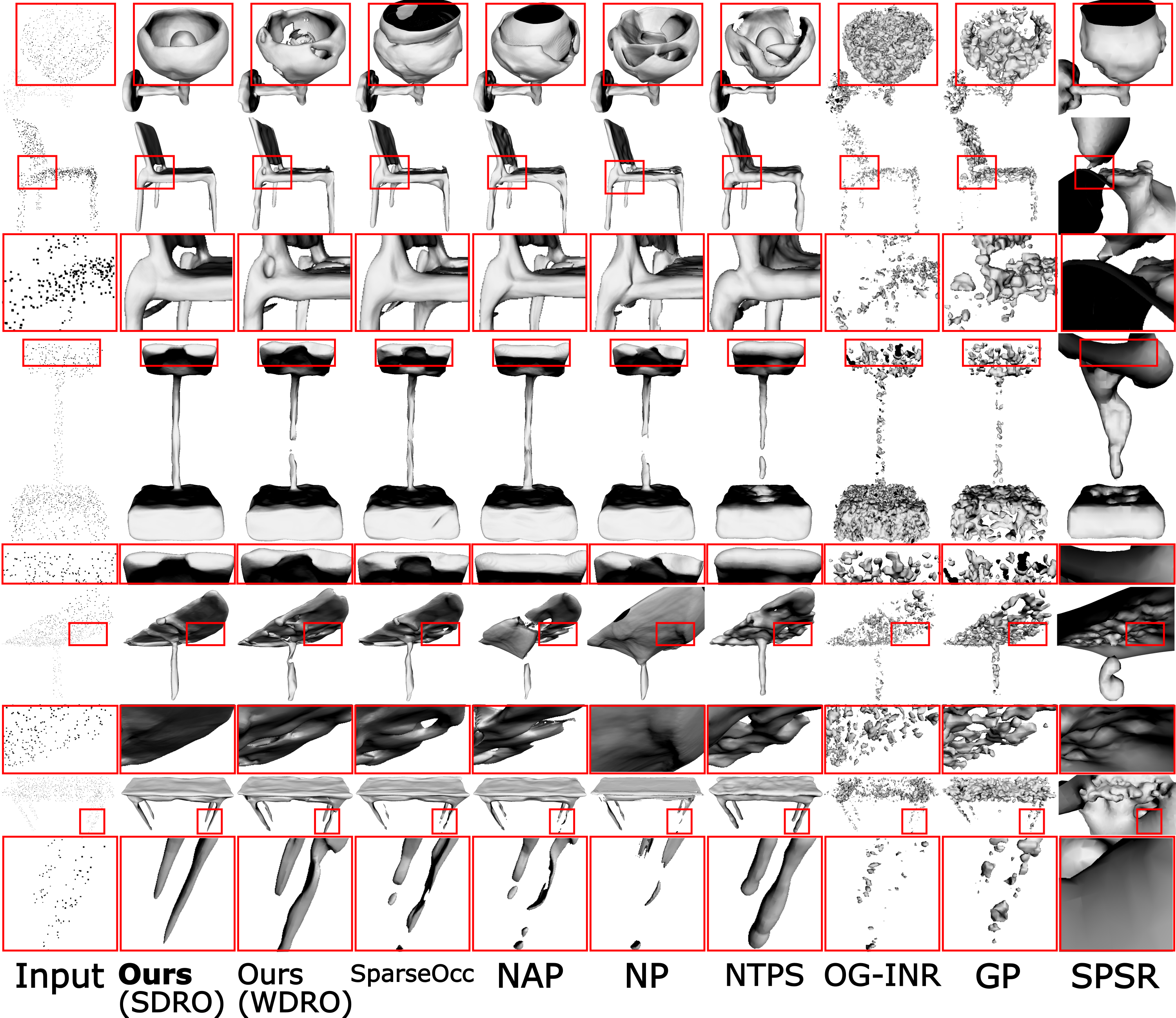}
\caption{We fit a neural SDF to a sparse noisy point cloud solely, using a distributionally robust loss function. Compared to the state-of-the-art, our method provides more faithful and robust reconstructions, as can be seen in these detailed and thin structures of ShapeNet \cite{shapenet} objects.} 
\label{fig:sn}
\end{figure}


3D reconstruction from point clouds remains a longstanding problem at the intersection of computer vision, graphics, and machine learning. While traditional optimization techniques like Poisson Reconstruction \citep{kazhdan2013screened,hou2022iterative} and Moving Least Squares \citep{guennebaud2007algebraic} perform well on dense, clean point clouds with accurate normal estimations, recent deep learning-based approaches offer improved robustness, especially when handling noisy or sparse inputs. These methods often eliminate the need for normal data. Many existing approaches rely on deep priors learned from large, fully labeled 3D datasets such as ShapeNet \citep{shapenet}, but this strategy requires expensive training, and the resulting models are still susceptible to generalization issues when exposed to out-of-distribution data—whether due to changes in input density or domain shifts, as noted by \citep{NeuralTPS,ouasfi2024robustifying}. Indeed, as demonstrated in \cref{tab:fs}, our unsupervised approach outperforms supervised generalizable models when tested on data that is sparser and diverges from the training data. This highlights the importance of developing learning frameworks that can ensure robust reconstruction under these challenging conditions.

Recent work \citep {nap} shows that strategies that can successfully recover SDF representations from dense point clouds, such as Neural-Pull (NP)  \citep{ma2020neural}, often struggle when the point cloud is sparse and noisy due to overfitting. As a consequence, the extracted shapes have missing parts and hallucinations  (\cf \cref{fig:3ds}, \cref{fig:sn}). Instead of relying on smoothness priors, \cite{nap} shifts the focus on how training distributions of spatial queries affect the performance of the SDF network. It introduces a special case of distributionally robust optimization (DRO) \citep{volpi2018generalizing, rahimian2019distributionally} for SDF learning. Within this framework, the network is trained by considering the worst-case distribution in terms of the loss function in a neighborhood around the observed training distribution. To tackle the challenging task of finding worst-case distributions, the training strategy proposed by \cite {nap} that we dub here NAP for Neural Adversarial Pull, relies on a first-order taylor approximation of the loss to find query-pointwise adversarial samples that are used to regularize the training. These are defined as query points that maximize the loss. In this paper, instead of relying on pointwise adversaries, we leverage recent advances in DRO literature to explore a tractable formulation for finding actual worst-case distributions for the first time in the context of reconstruction from point cloud. Our solution also proves to be more resilient to noise in the sparse input setting.




One key design choice that determines the type and the level of input noise that can be mitigated as well as the tractability of the problem is the uncertainty set. That is the set of distributions where the worst-case distribution can be found. This is usually defined as a neighborhood of the initial training distribution. To measure the distance between distributions, various metrics have been explored in DRO literature, including f-divergence \citep{ben2013robust, miyato2015distributional, namkoong2016stochastic}, alongside the Wasserstein distance \citep{wdro,kuhn_dro}. The latter has demonstrated notable advantages in terms of efficiency and simplicity, in addition to being widely adopted in computer vision and graphics downstream applications \citep{rubner2000earth, pele2008linear, solomon2015convolutional, solomon2014earth}, as it takes into account the geometry of the sample space, in contrast to other metrics.

In order to learn a neural SDF from a sparse noisy point cloud within a DRO framework, we proceed in this work as follows. $\bullet$ We first present a tractable implementation for this problem (SDF WDRO) benefiting from the dual reformulation \citep{wdro} of the DRO problem with the Wasserstein distribution metric \citep{kuhn_dro,  wdro, wrm, udr}. We build on NP \citep{ma2020neural}, but instead of using their predefined empirical spatial query distribution (sampling normally around each of the input points), we rely on queries from the worst-case distribution in the Wasserstein ball around the empirical distribution. While this reduces overfitting and leads to more robust reconstructions thanks to using more informative samples throughout training instead of overfitting on easy ones, this improvement comes at the cost of additional training time compared to the NP baseline as shown in \cref{fig:abl_time}. $\bullet$ Furthermore, by interpreting the Wasserstein distance computation as a mass transportation problem, recent advances in Optimal Transport shows that it is possible to obtain theoretically grounded approximations by regularizing the original mass transportation problem with a relative entropy penalty on the transport plan (\eg \cite{cuturi2013sinkhorn}). The resulting distance is referred to as Sinkhorn distance. Thus, we show subsequently that substituting the Wasserstein distance with the Sinkhorn one in our SDF DRO problem results in a computationally efficient dual formulation \citep{sdro} that significantly improves the convergence time of our first baseline SDF WDRO. The training algorithm of the resulting SDF SDRO is outlined in \cref{alg:sdro}. Thanks to the entropic regularization, SDF SDRO produces more diffused spatial adversaries by smoothing the worst-case distribution \citep{sdro,sdro_general, Blanchet_2020}. As a result, errors in the SDF approximation are better distributed across the shape, improving overall performance. 

Through extensive quantitative and qualitative evaluation under several real and synthetic benchmarks for object, non rigid and scene level shape reconstruction, our results show that our final method (SDF SDRO) outperforms SDF WDRO, the baseline NP, as well as the most relevant competition, notably the current state-of-the-art in unsupervised learning of SDFs from sparse point cloud, such as NTPS \citep{NeuralTPS}, NAP \citep{nap} and SparseOcc \cite{sparseocc}. 

\noindent\textbf{Summary of intuition and contribution}
We understand the approach proposed in \cite{nap} as a means of distributing the SDF approximation errors evenly throughout the 3D shape. 
Point cloud low-density and noisy areas are where this SDF error tends to concentrate. While in \cite{nap} the query points are independently perturbed within a local radius, our key idea is to construct a distribution of the most challenging query samples around the shape in terms of the loss function by “perturbing” the initial distribution of query points. The cost of this perturbation is controlled globally through an optimal transport distance. Minimizing the expected loss over this distribution flattens the landscape of the loss spatially, ensuring that the implicit model behaves consistently in the 3D space. As demonstrated by \cite{staib2017distributionally}, not only does this generalize the approach proposed in \cite{nap}, but also provides stronger adversaries that can be used to regularize the training, which justifies our superior results.

\section{Related Work}
\label{sec:rel}
\vspace{-5pt}

\noindent\textbf{Reconstruction from Point Clouds}
Traditional methods for reconstructing shapes include combinatorical techniques that divide the input point cloud into parts, such as using alpha shapes \citep{bernardini1999ball}, Voronoi diagrams \citep{amenta2001power}, or triangulation \citep{cazals2006delaunay,liu2020meshing,rakotosaona2021differentiable}. An alternative approach, is to define implicit functions, whose zero level set represents the target shape, using the input samples. This can be achieved by incorporating global smoothing priors \citep{williams2022neural,lin2022surface,williams2021neural,ouasfi2024robustifying}, such as radial basis functions \citep{carr2001reconstruction} and Gaussian kernel fitting \citep{scholkopf2004kernel}, or local smoothing priors like moving least squares \citep{mercier2022moving,guennebaud2007algebraic,kolluri2008provably,liu2021deep}. A different approach consists of solving a Poisson equation with boundary conditions \citep{kazhdan2013screened}. In recent years, there has been a shift towards representing these implicit functions using deep neural networks, with parameters learned via gradient descent, either in a supervised  (\eg \citep{boulch2022poco,williams2022neural,huang2023neural,peng2020convolutional,chibane2020implicit,lionar2021dynamic,ouasfi2023mixing,peng2021shape,ouasfi2022few}) or unsupervised manner.
These implicit representations \cite{mescheder2019occupancy,park2019deepsdf} aleviate many of the shortcomings of explicit ones
(\eg meshes \cite{wang2018pixel2mesh,kato2018neural,jena2022neural} and point clouds \cite{fan2017point,aliev2020neural,kerbl20233d})
in modelling shape, radiance and light fields (\eg \cite{mildenhall2020nerf,yariv2021volume,wang2021neus,jain2021dreamfields,chan2022efficient,li2023learning,li2023regularizing,jena2024geo}), as they allow to model arbitrary topologies at virtually infinite resolution. 

\noindent\textbf{Unsupervised Implicit Neural Reconstruction} A neural network is used to fit a single point cloud without additional supervision in this setting. Regularizations, such as the spatial gradient constraint based on the Eikonal equation proposed by Gropp et al. \cite{gropp2020implicit}, the spatial Laplacian constraint introduced in \cite{ben2022digs}, and Lipschitz regularization on the network \citep{liu2022learning} are used to constrain the learned SDF, leading to performance improvement. \cite{sitzmann2020implicit} introduced periodic activations. \cite{lipman2021phase} shows that an occupancy function can be learned such that its log transform converges to a distance function. \cite{atzmon2020sal} learns SDF from unsigned distances, with normal supervision on the spatial gradient of the function \citep{atzmon2020sald}. \cite{ma2020neural} express the nearest point on the surface as a function of the neural signed distance and its gradient. Several state-of-the-art reconstruction methods (\eg \cite{younes2025sparsecraft,nap,huang2023neusurf, NeuralTPS,ma2022surface,ma2022reconstructing,LPI,ma2020neural}) build on this strategy. Self-supervised local priors are used to handle very sparse inputs \cite{ma2022reconstructing} or enhance generalization \cite{ma2022surface}. \cite{boulch2021needrop} proposed to learn an occupancy function by assuming that needle end points near the surface are statistically located on opposite sides of the surface. \cite{williams2021neural} solved a kernel ridge regression problem using points and their normals. \cite{peng2021shape} proposed a differentiable Poisson solving layer to efficiently obtain an indicator function grid from predicted normals. \cite{koneputugodage2023octree} learns an implicit field using Octree-based labeling as a guiding mechanism. \cite{NeuralTPS} provides additional coarse surface supervision to the shape network using a learned surface parametrization. However, when the input is sparse and noisy, most of the methods mentioned above continue to face challenges in generating accurate reconstructions due to insufficient supervision. \cite{sparseocc} learns an occupancy function by sampling from its uncertainty field and stabilizes the optimization by biasing the occupancy function towards minimal entropy fields. \cite{nap} augments the training with adversarial samples around the input point cloud. Differently from this literature, we explore here a new paradigm for learning unsupervised neural SDFs for the first time, namely through DRO with Wasserstein uncertainty sets.






\begin{figure}[t]
\centering
\vspace{-5pt}
\includegraphics[width=0.7\linewidth]{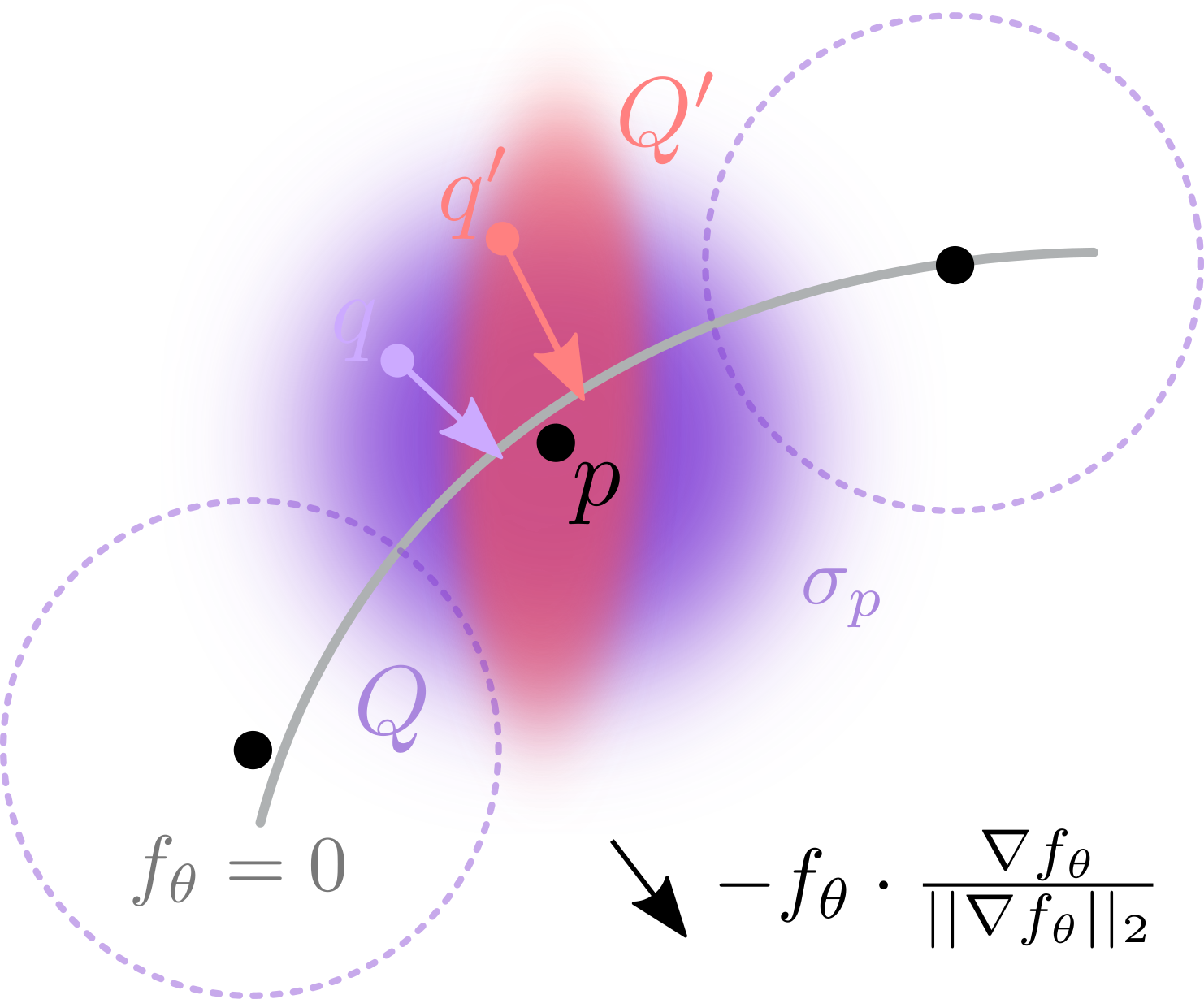}
\caption{\footnotesize We learn a neural SDF $f_{\theta}$ from a point cloud (black dots) by minimizing the error between projection of spatial queries $\{q\}$ on the level set of the field (gray curve) and their nearest input point $p$. Instead of learning with a standard predefined distribution of queries $Q$, we optimize for the worst-case query distribution $Q'$ within a ball of distributions around $Q$.}
\vspace{-15pt}
\label{fig:overview}
\end{figure}

\section{Method}
\vspace{-5pt}
Let   $\Xi$  be a subset of   ${\mathbb R}^{3}$, and let  $\mathcal{M}(\Xi)$, and $\mathcal{P}(\Xi)$  represent the set of measures and the set of probability measures on  $\Xi$, respectively.
Given a noisy, sparse unoriented point cloud $\mathbf{P}\subset{\Xi}^{ N_p}$, our objective is to  reconstruct a corresponding watertight 3D shape reconstruction, \ie the shape surface  $\mathcal{S}$ that best fits the point cloud  $\mathbf{P}$. 
To achieve this, we parameterise the  shape function $f$  to be learned with an MLP $f_\theta$ that implicitly represents the signed distance field  to the target shape $\mathcal{S}$. 
The reconstructed shape $\hat{\mathcal{S}}$ is represented as  the zero level set of the SDF (signed distance function) $f_\theta$: $
\hat{\mathcal{S}} = \{ q\in\mathbb{R}^3 \mid f_\theta(q) = 0\}
$. In practice, we use  the Marching Cubes algorithm \citep{lorensen1987marching} to extract an explicit triangle mesh for $\hat{\mathcal{S}}$  by querying neural network $f_\theta$. 

\subsection{Learning an SDF by Query Neural Pulling}



Neural Pull (NP) \citep{ma2020neural} approximates a signed distance function by pulling query points to their their nearest input point cloud sample using the gradient of the SDF network. The normalized gradient is multiplied by the  negated signed distance predicted by the network in order to pull both inside and outside queries to the surface. Query points are  drawn from normal distributions centered at at input samples $\{p\}$, with  local standard deviations $\{\sigma_p\}$ defined  as the maximum euclidean distance to the $K$ nearest  points to $p$ in $\mathbf{P}$:
\begin{equation}
\mathfrak{Q} := \bigcup_{p\in \mathbf{P}}\{q \sim \mathcal{N}(p,\sigma_p \mathbf{I}_3)\},
\label{equ:sample}
\end{equation} 
The  neural SDF $f_\theta$ is trained in \cite{ma2020neural} with empirical risk minimization (ERM) using the following objective:
\begin{equation}
\mathcal{L}(\theta,q) = ||q - f_\theta(q) \cdot \frac{\nabla f_\theta(q)}{||\nabla f_\theta(q)||_2} - p||_2^2, \ 
\label{equ:np}
\end{equation}

where $p$ is the closest point to $q$ in  $\mathbf{P}$.  By minimizing  the expected loss under the empirical distribution $Q = \sum _{q \in \mathfrak{Q}} \delta_q$  owhere  $\delta_q$ is the dirac distribution or the unit mass on $q$, this objective ensures that the samples in $\mathbf{P}$ are on  zero level set of the neural SDF $f_\theta$.


\subsection{Neural SDF DRO}
\vspace{-5pt}
Inspired by \cite{nap}, we focus on how to distribute the SDF approximation errors evenly throughout the shape as  without regularization these errors tends to concentrate in low-density and noisy areas. NAP \cite{nap} introduces the following regularization term: 
\begin{equation}
\mathcal{L}_{\text{NAP}}(\theta,Q) = \underset{q\sim Q}{\mathbb{E}} \max_{\delta,\, ||\delta||_2<\rho} \mathcal{L}(\theta,q+\delta),
\label{equ:nap}
\end{equation}
Where the perturbation radius $\rho$ controles the distance to the spatial adversaries. Using a first order Taylor expansion of the loss, this  problem is solved efficiently in \cite{nap} by deriving individual perturbations on the query points $q$ as follows: 
\begin{equation}
\hat{\delta} = \rho  \,\frac{\nabla_q\mathcal{L}(\theta,q)}{||\nabla_q\mathcal{L}(\theta,q)||_2}
\label{equ:delta}
\end{equation}

We consider the DRO problem introduced by NAP with Wasserstein uncertainty sets ( \cref{equ:dro}).  We optimize the parameters of the SDF network  $\theta$ under the worst-case expected loss among a ball of distributions  $Q'$  in this uncertainty set \citep{wdro_general, wdro},: 


\begin{equation}
\begin{split}
\inf_\theta \sup_{Q' : \mathcal{W}_c(Q', Q) < \epsilon} 
&\underset{q' \sim Q'}{\mathbb{E}} \, \mathcal{L}(\theta, q'), \\
\text{where} \quad \mathcal{W}_c(Q', Q) &:= \inf_{\gamma \in \Gamma(Q', Q)} \int c \, d\gamma.
\end{split}
\label{equ:dro}
\end{equation}

Here, $\epsilon>0$ and $\mathcal{W}_c$ denotes the optimal transport (OT) or a Wasserstein distance for a cost function $c$, defined as the infimum over the set $\Gamma({Q'}, Q)$  of couplings whose marginals are $Q'$ and $Q$. We refer the reader to the body of work in \eg \cite{wdro_general, wdro} for more background.

\noindent\textbf{Neural SDF Wasserstein DRO (WDRO)}
A tractable reformulation of the optimization problem defined in Equation  \cref{equ:dro} is made possible thanks to the following duality result \citep{wdro}. For upper semi-continuous loss functions and non-negative lower semi-continuous costs satisfying $c\left(z, z^{\prime}\right)=0$ iff $z=z^{\prime}$,  the optimization problem (\cref{equ:dro})  is equivalent to: 


\begin{align}
\inf_{\theta, \lambda \geq 0} &\left\{ \lambda \epsilon + \mathcal{L}_{\text{WDRO}}(\theta, Q) \right\}, \notag \\
\text{where}\, \mathcal{L}_{\text{WDRO}}(\theta, Q) &= \mathbb{E}_{q \sim Q} 
\left[\sup_{q'} \left\{ \mathcal{L}(\theta, q') - \lambda c(q', q) \right\} \right].
\label{equ:wdro}
\end{align}

As shown in \cite{udr}, solving the optimization above with a fixed dual variable $\lambda$ yields inferior results to the case where $\lambda$ is updated. In fact, optimizing $\lambda$ allows to capture global information when solving the outer minimization, whilst only local information (local worst-case spatial queries) is considered when minimizing $\mathcal{L}_{\text{WDRO}}$ solely. 


Following \cite{udr}, the optimization in Equation \cref{equ:wdro} can be carried as follows: Given the current model parameters $\theta$ and the dual variable $\lambda$, the worst-case spatial query $q'$ corresponding to a query $q$ drawn from the empirical distribution $Q$ can be obtained through a perturbation of $q$ followed by a few steps of iterative gradient ascent over $\mathcal{L}(\theta,q^{\prime})-\lambda c\left(q^{\prime}, q\right)$. Subsequently, inspired by the Danskin’s theorem,    $\lambda$ can be updated  accordingly $\lambda \leftarrow \lambda-\eta_\lambda\left(\epsilon-\frac{1}{N_b} \sum_{i=1}^{N_b} c\left(q_i^{\prime}, q_i\right)\right)$, where $N_b$ represents the query batch size, and $\eta_\lambda>0$ symbolizes a learning rate \cite{udr} . The current batch loss $\mathcal{L}_{\text{WDRO}}$ can then be backpropagated. We provide an Algorithm in supplemental material recapitulating this training. 




While NAP consists of a hard-ball projection with locally adaptive radii, WDRO samples from the worst case distribution around the shape, (Equation \cref{equ:wdro}) through  a soft-ball projection controlled by the parameter $\lambda$ that is adjusted throughout the training. The $\lambda$ update rule ensures that it grows when the worst-case sample distance from the initial queries exceeds the Wasserstein ball radius $\epsilon$. While this  approach provides promising results, it suffers from rather slow convergence, as shown in Figure \cref{fig:abl_time}. Furthermore, because our nominal distribution $Q$ is finitely supported, the worst-case distribution generated with WDRO is proven to be a discrete distribution \citep{wdro_general}, even while the underlying actual distribution is continuous. As pointed out in \cite{sdro}, this questions  whether WDRO hedges the right family of distributions or  generates too conservative solutions. In the next section, we show how these limitations can be addressed by taking inspiration from recent advances in Optimal Transport.  

\noindent\textbf{Neural SDF Wasserstein DRO with entropic regularization (SDRO)} One key technical aspect underpinning the recent achievements of Optimal Transport  in various applications lies in the use of regularization, particularly entropic regularization \citep{sdro_general}. This approach has paved the way for efficient computational methodologies ( \eg \cite{cuturi2013sinkhorn}) to obtain theoretically-grounded approximations of Wasserstein distances. Building upon these advancements, recent work \citep{sdro_general, sdro} extend the framework of Wasserstein Distributionally Robust Optimization with entropic regularization by substituting the Wasserstein distance in Equation \cref{equ:dro} with the Sinkhorn distance \citep{sdro}. 

For $P,Q \in \mathcal{P}(\Xi)$,  the Sinkhorn distance is defined as:
\begin{equation}
\mathcal{W}_\rho(P,Q)=\inf _{\gamma \in \Gamma(P,Q)}\left\{\mathbb{E}_{(x, y) \sim \gamma}[c(x, y)]+\rho H(\gamma \mid \mu \otimes \nu)\right\},
\end{equation}
where $\rho \geq 0$ is a regularization parameter.  $\mu$ and  $\nu$ are two reference measures in $\mathcal{M}(\Xi)$ such that $P$ and $Q$ are absolutely continuous \wrt to $\mu$ and $\nu$ respectively, $H(\gamma \mid \mu \otimes \nu)$ denotes the relative entropy of $\gamma$ with respect to the product measure $\mu \otimes \nu$ :
\begin{equation}
H(\gamma \mid \mu \otimes \nu)=\mathbb{E}_{(x, y) \sim \gamma}\left[\log \left(\frac{\mathrm{d} \gamma(x, y)}{\mathrm{d} \mu(x) \mathrm{d} \nu(y)}\right)\right],
\end{equation}
where $\frac{\mathrm{d} \gamma(x, y)}{\mathrm{d} \mu(x) \mathrm{d} \nu(y)}$ stands for the density ratio of $\gamma$ with respect to $\mu \otimes \nu$ evaluated at $(x, y)$. 

Compared to the  Wasserstein distance, Sinkhorn distance   regularizes the original mass transportation problem with relative entropy penalty on the transport plan. The choice of the reference measures $\mu$ and $\nu$ acts as a prior on the DRO problem. Following \cite{sdro}, we fix $\mu$ as our empirical distribution $Q$ and $\nu$ as the Lebesgue measure. Consequently, optimization problem in Equation \cref{equ:dro} with the Sinkhorn distance admits the following dual form: 
\begin{equation}
\inf _{\theta, \lambda \geq 0}\left\{\lambda \bar{\epsilon}+\lambda \rho \mathbb{E}_{q \sim Q }\left[\log \mathbb{E}_{q^{\prime} \sim \mathbb{Q}_{q, \rho}}\left[e^{\mathcal{L}(\theta,q^{\prime}) /(\lambda \rho)}\right]\right]\right\},
\label{equ:sdro1}
\end{equation}
where $\bar{\epsilon}$ is a constant that depends on $\rho$ and $\epsilon$ (\cite{sdro}). Additionally, distribution $\mathbb{Q}_{q, \rho}$ is defined through:
\begin{equation}
\mathrm{d} \mathbb{Q}_{x, \rho}(z):=\frac{e^{-c(x, z) / \rho}}{\mathbb{E}_{u \sim \nu}\left[e^{-c(x, u) / \rho}\right]} \mathrm{d} \nu(z).
\label{equ:samp_sdro}
\end{equation}

As discussed in \cite{sdro}, optimizing $\lambda$ within problem \cref{equ:sdro1} leads to instability. Hence, for a given fixed  $\lambda > 0$, optimization \cref{equ:sdro1} can be carried practically by sampling a set of $N_s$ samples $q^{\prime}\sim \mathbb{Q}_{q, \rho}$ for each query $q$, then backpropagating the following distributionaly robust loss:
\begin{equation}
    \mathcal{L}_{\text{SDRO}}(\theta,Q) = \lambda \rho \mathbb{E}_{q \sim Q }\left[\log \mathbb{E}_{q^{\prime} \sim \mathbb{Q}_{q, \rho}}\left[e^{\mathcal{L}(\theta,q^{\prime}) /(\lambda \rho)}\right]\right].
    \label{equ:sdro}  
\end{equation}
 \cref{alg:sdro} summarizes the training of our SDRO based method.  



\subsection{Training Objective}
Similar to \cite{nap} we train using the strategy of \cite{liebel2018auxiliary} which combines  the  original objective and the distributionally robust one:


\begin{equation}
\begin{split}
\mathfrak{L}(\theta,q) = \frac{1}{2\lambda_1}\mathcal{L}(\theta,q) + \frac{1}{2\lambda_2}\mathcal{L}_{\text{DRO}}(\theta,q) \\
+ \ln(1+\lambda_1) + \ln(1+\lambda_2).
\end{split}
\label{equ:final}
\end{equation}

where $\lambda_1$ and  $\lambda_2$ are learnable weights and $\mathcal{L}_{\text{DRO}}$ is either $\mathcal{L}_{\text{SDRO}}$  or $\mathcal{L}_{\text{WDRO}}$. Our training procedure is shown in Algorithms \ref{alg:wdro} and \cref{alg:sdro}. 

\begin{algorithm}[h!]
\small
\begin{algorithmic}
\Require Point cloud $\textbf{P}$, learning rate $\alpha$, number of iterations $N_{\text{it}}$, batch size $N_b$.\\
SDRO hyperparameters: $\rho$, $\lambda$, $N_s$.
\Ensure Optimal parameters ${\theta}^*$.
\State Compute local st. devs. $\{\sigma_p\}$ ($\sigma_p=\max_{t\in K\text{nn}(p,\mathbf{P})}||t-p||_2$).
\State $ \mathfrak{Q} \leftarrow$ sample($\textbf{P}$,$\{\sigma_p\}$) (Equ. \cref{equ:sample})
\State Compute nearest points in $\textbf{P}$ for all samples in $\in \mathfrak{Q}$. 
\State Initialize $\lambda_1 = \lambda_2 = 1$.  
\For{$N_{\text{it}}$ times}
\State Sample $N_b$  query points   $\{  q, q\sim Q    \}$.
\State For each $q$, sample $N_s$ points $\{ q^{\prime}, q^{\prime}\sim \mathbb{Q}_{q, \rho} \}$. (Equ.\cref{equ:samp_sdro})
\State Compute SDRO losses $\{\mathcal{L}_{\text{SDRO}}(\theta,q)\}$ (Equ. \cref{equ:sdro})
\State Compute combined losses $\{\mathfrak{L}(\theta,q)\}$ (Equ. \cref{equ:final})
\State $(\theta, \lambda_1, \lambda_2) \leftarrow (\theta, \lambda_1, \lambda_2) - \alpha \nabla_{\theta,\lambda_1,\lambda_2} \Sigma_q \mathfrak{L}(\theta ,q)$
\EndFor
\end{algorithmic}
\caption{The training procedure of our method with SDRO.}
\label{alg:sdro}
\end{algorithm}
\vspace{-10pt}

\section{Results}
To assess the performance of our approach, we conducted evaluations on widely used 3D reconstruction benchmarks. In line with previous research, we assess the accuracy of 3D meshes generated by our MLPs after convergence. We benchmark our method against SOTA methods for sparse unsupervised reconstruction, including NP~\citep{ma2020neural}, NAP~\citep{nap}, SparseOcc~\citep{sparseocc}, and NTPS~\citep{NeuralTPS}. 
Further, we extend our comparisons to include approaches such as SAP~\citep{peng2021shape}, DIGS~\citep{ben2022digs}, NDrop~\citep{boulch2021needrop}, and NSpline~\citep{williams2021neural}, as well as hybrid methods that incorporate both explicit and implicit representations, such as OG-INR~\citep{koneputugodage2023octree} and GridPull (GP)~\citep{chen2023gridpull}. 
We also evaluate our approach relative to SOTA supervised methods. These include robust, generalizable feed-forward methods such as POCO~\citep{boulch2022poco}, CONet~\citep{peng2020convolutional}, and NKSR~\citep{Huang_2023_CVPR}, in addition to prior-based optimization strategies tailored for sparse data, like On-Surf~\citep{ma2022reconstructing}. Following the settings adopted by NAP, our experiments employ point clouds with $N_p = 1024$ points.

\subsection{Metrics}
\vspace{-5pt}

We evaluate our method using standard metrics commonly employed for 3D reconstruction tasks. Specifically, we compute the L1 \textbf{Chamfer Distance} (CD$_1$) and L2 \textbf{Chamfer Distance} (CD$_2$), both scaled by a factor of $10^2$. Additionally, we compute the \textbf{F-Score (FS)}, based on Euclidean distance, and the \textbf{Normal Consistency (NC)} between the meshes generated by our approach and the ground-truth. Detailed mathematical formulations for these metrics are provided in the supplementary material.

\vspace{-5pt}
\subsection{Datasets and input definitions}
\vspace{-5pt}

We evaluate our approach on several benchmark datasets representing a variety of 3D data. 

\textbf{ShapeNet}~\citep{shapenet} provides a diverse set of synthetic 3D models across $13$ distinct categories. In line with previous work, we report results on the Table, Chair, and Lamp classes, utilizing the train/test splits specified in \cite{williams2021neural}. For each mesh, we generate noisy input point clouds by sampling 1024 points and adding Gaussian noise with a standard deviation of $0.005$, as done in \cite{boulch2022poco, peng2020convolutional, nap}.
\textbf{Faust}~\citep{Bogo:CVPR:2014} includes real 3D scans of $10$ different human body identities, each captured in $10$ distinct poses. We sample $1024$ points from these scans to serve as input for our method.
\textbf{3D Scene}~\citep{zhou2013dense} contains large-scale, real-world scenes acquired with a handheld commodity range sensor. We follow the protocols in \cite{NeuralTPS, jiang2020local, ma2020neural, nap} to generate sparse point clouds with a density of $100$ points per m$^3$ and present results for several scenes, including Burghers, Copyroom, Lounge, Stonewall, and Totempole.
\textbf{SemanticPOSS}~\cite{semantic} contains LiDAR data collected from $6$ sequences of road scenes. Each scan captures a $51.2$m range ahead, $25.6$m on either side, and $6.4$m vertically. We provide qualitative results from each of these sequences.
Finally, we extend our evaluation to challenging scenes from the \textbf{BlendedMVS}~\citep{bmvs} dataset, which is a multi-view stereo dataset, with scenes consisting of architecture, sculptures, and small objects with complex backgrounds as well as \textbf{Tanks and Temples}~\citep{tanks} dataset, which consists of large-scale indoor and outdoor scenes, with high-resolution images captured by a handheld monocular RGB camera. For both the datasets, sparse views are used with VGGSfM \citep{wang2024vggsfm} to generate sparse, noisy input point clouds for our setup.




\begin{figure}[!t]
\centering
\includegraphics[width=1.0\linewidth]{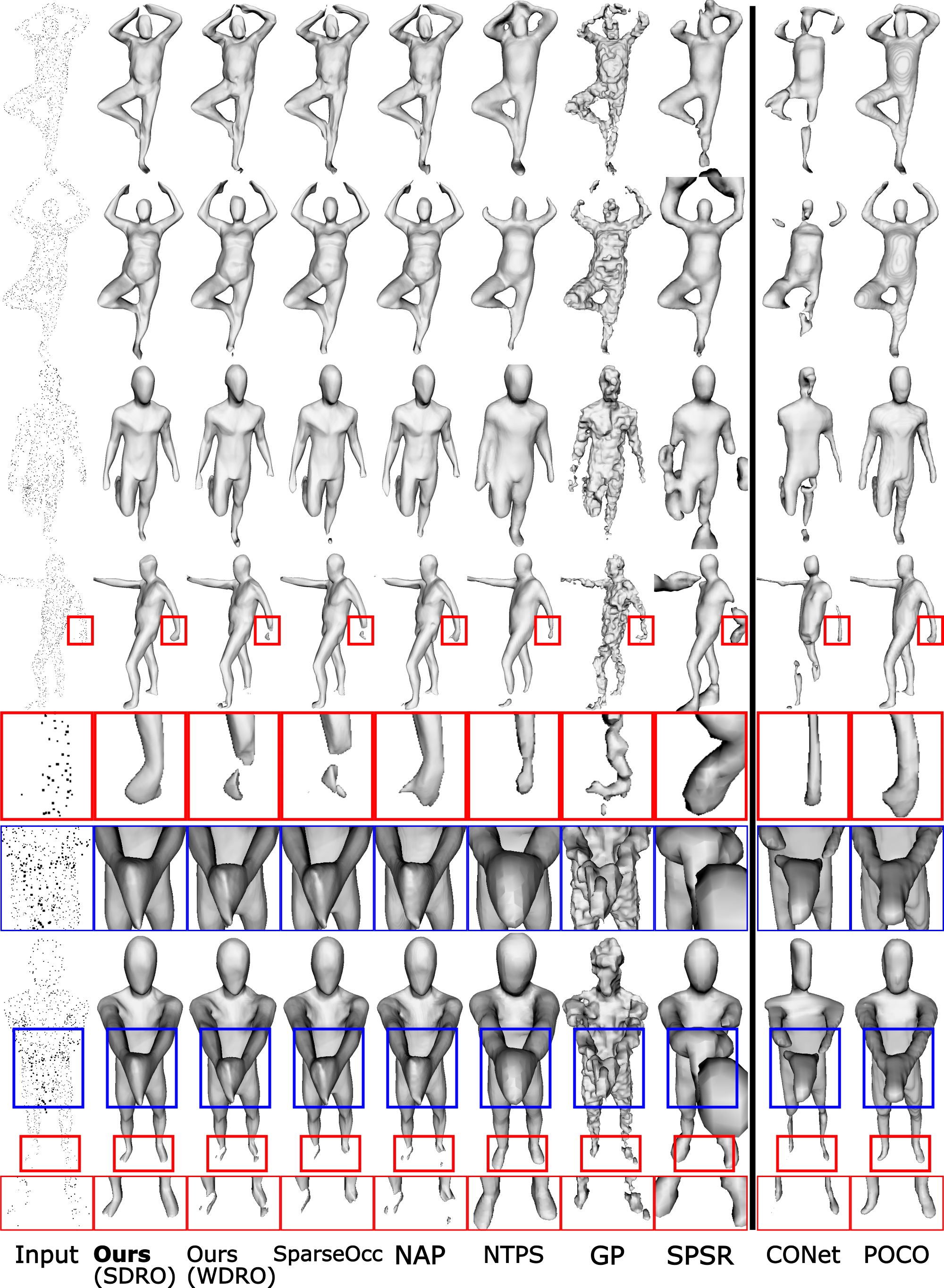}
\vspace{-5pt}
\caption{Faust \cite{Bogo:CVPR:2014} reconstructions. \textbf{CONet and POCO use data priors}.} 
\label{fig:fs}
\end{figure}



\vspace{-5pt}
\subsection{Implementation details}
\label{sec:implementation}
\vspace{-5pt}





Our MLP model, denoted as $f_\theta$, follows the architecture outlined in Neural Pull (NP)~\citep{ma2020neural}. We train the model using the Adam optimizer with a batch size of $N_b = 5000$ and set $K = 51$ to compute the local standard deviations $\sigma_p$, in line with NP. Training is performed on a single NVIDIA RTX A6000 GPU. For fair and practical comparison, we select the optimal evaluation epoch for each method based on Chamfer distance between the reconstructed and input point clouds, choosing the epoch that minimizes this metric. We also conduct a hyperparameter search on the SRB benchmark to determine the best parameters for our method.

For the Wasserstein Robust DRO (WRDO) approach, we perform two gradient ascent steps ($N_{it}^{wdro} = 2$) with a learning rate $\alpha_{wdro} = 10^{-3}$ in the inner loop. The dual variable is initialized as $\lambda = 80$, and the Wasserstein ball radius is fixed at $\epsilon = 10^{-4}$. For Standard DRO (SDRO), we use $N_s = 5$ samples per query point $q \sim Q$, with $\lambda = 20$ in our experiments. The transport cost is defined as $c(\cdot, \cdot) = \frac{1}{2} || \cdot - \cdot ||^2$, implying that sampling from $\mathbb{Q}_{q, \rho}$ follows a Gaussian distribution $\mathcal{N}(q, \rho \mathbf{I}_3)$.

\vspace{-5pt}
\subsection{Object level reconstruction}
\vspace{-5pt}
We evaluate the reconstruction of ShapeNet~\citep{shapenet} objects from sparse and noisy point clouds. A quantitative comparison is presented in ~\cref{tab:sn}, while ~\cref{fig:sn} offers a qualitative assessment of the results. Our approach, based on Wasserstein Robust DRO (WDRO), outperforms existing methods in terms of reconstruction accuracy, as measured by CD$_1$ and CD$_2$. When combined with the SDRO loss, our method further improves across all evaluation metrics. This is reflected in the visually enhanced reconstruction quality, which demonstrates superior detail and structure preservation. While NTPS produces generally acceptable coarse reconstructions, its use of thin plate spline smoothing limits its ability to capture finer details. NAP and SparaseOcc are able to produce better reconstructions but struggle under high levels of noise. Additionally, we find that OG-INR struggles to achieve satisfactory convergence under sparse and noisy conditions, despite its success in denser scenarios aided by Octree-based sign fields.


\begin{table}[t!]
\begin{minipage}{0.48\textwidth}
\centering
\scalebox{0.7}{
\begin{tabular}{lllll}
\hline
       & \textbf{CD1}     & \textbf{CD2}    & \textbf{NC}       & \textbf{FS}       \\ \hline
SPSR \cite{kazhdan2013screened} & 2.34          & 0.224          & 0.74          & 0.50         \\
OG-INR \cite{koneputugodage2023octree} & 1.36          & 0.051          & 0.55          & 0.55          \\
NP \cite{ma2020neural} & 1.16          & 0.074          & 0.84          & 0.75          \\
GP \cite{chen2023gridpull} &1.07	&0.032	& 0.70 & 0.74\\
NTPS \cite{NeuralTPS} & 1.11          & 0.067          & \cellcolor{tabsecond}0.88& \cellcolor{tabthird}0.74         \\ 
NAP \cite{nap} & \cellcolor{tabsecond}0.76 & \cellcolor{tabthird}0.020 & \cellcolor{tabthird}0.87         & \cellcolor{tabsecond}0.83 \\
SparseOcc \cite{sparseocc} & \cellcolor{tabsecond}0.76& \cellcolor{tabthird}0.020 & \cellcolor{tabsecond}0.88     & \cellcolor{tabsecond}0.83 \\
Ours (WDRO)& \cellcolor{tabthird}0.77& \cellcolor{tabsecond}0.015& \cellcolor{tabthird}0.87&\cellcolor{tabsecond}0.83 \\ 
 \textbf{Ours} (SDRO)&  \cellcolor{tabfirst}0.63&  \cellcolor{tabfirst}0.012& \cellcolor{tabfirst}0.90&\cellcolor{tabfirst}0.86\\
\hline 
\end{tabular}}
\caption{ShapeNet \citep{shapenet} reconstructions from sparse noisy unoriented point clouds.}
\label{tab:sn}
\end{minipage}
\hfill
\begin{minipage}{0.48\textwidth}
\centering
\scalebox{0.7}{
\begin{tabular}{lllll}
\hline
               & \textbf{CD1}     & \textbf{CD2} & \textbf{NC}       & \textbf{FS} \\ \hline

POCO \cite{boulch2022poco} & 0.308     & 0.002   & 0.934        & \cellcolor{tabfirst} 0.981    \\
CONet \cite{peng2020convolutional} & 1.260     & 0.048    & 0.829      & 0.599    \\
On-Surf \cite{ma2022reconstructing} &0.584    &0.012   &0.936 & 0.915  \\
NKSR \cite{Huang_2023_CVPR} &0.274    &  \cellcolor{tabsecond}0.002   &  0.945 & \cellcolor{tabfirst}0.981  \\
\hline
SPSR \cite{kazhdan2013screened} & 0.751     & 0.028    & 0.871        & 0.839    \\
GP \cite{chen2023gridpull} & 0.495    & 0.005   &0.887 &  0.945\\
NTPS \cite{NeuralTPS} & 0.737     & 0.015     & 0.943          & 0.844     \\
NAP \cite{nap} & \cellcolor{tabfirst}0.220   & \cellcolor{tabfirst}0.001    & \cellcolor{tabfirst}0.956 & \cellcolor{tabfirst}0.981    \\
SparseOcc \cite{sparseocc} &0.260    & \cellcolor{tabsecond}0.002     & 0.952 & 0.974    \\ 
Ours (WDRO)& \cellcolor{tabthird}0.255& \cellcolor{tabsecond}0.002&\cellcolor{tabthird}0.953& \cellcolor{tabthird}0.977\\ 
 \textbf{Ours} (SDRO)& \cellcolor{tabsecond}0.251&\cellcolor{tabsecond}0.002& \cellcolor{tabsecond}0.955&\cellcolor{tabsecond}0.979\\  
\hline
\end{tabular}}
\label{tab:fs}
\caption{Faust \citep{Bogo:CVPR:2014} reconstructions from sparse noisy unoriented point clouds. \textbf{POCO, CONet, On-Surf and NKSR use data priors.}}
\label{tab:fs}
\end{minipage}

\vspace{-5pt}
\end{table}

\begin{table*}[t!]
\centering
\scalebox{0.65}{
\begin{tabular}{l|lllllllllllllllllll}
\cline{1-19}
\multicolumn{1}{c|}{} & \multicolumn{3}{c|}{\textbf{Burghers}}                                                       & \multicolumn{3}{c|}{\textbf{Copyroom}}                                                       & \multicolumn{3}{c|}{\textbf{Lounge}}                                                         & \multicolumn{3}{c|}{\textbf{Stonewall}}                                                      & \multicolumn{3}{c|}{\textbf{Totemple}}                                                       & \multicolumn{3}{c}{\textbf{Mean}}                                    &  \\ 
\multicolumn{1}{c|}{}                  & \multicolumn{1}{l}{CD1}  & \multicolumn{1}{l}{CD2}  & \multicolumn{1}{l|}{NC}    & \multicolumn{1}{l}{CD1}  & \multicolumn{1}{l}{CD2}  & \multicolumn{1}{l|}{NC}    & \multicolumn{1}{l}{CD1}  & \multicolumn{1}{l}{CD2}  & \multicolumn{1}{l|}{NC}    & \multicolumn{1}{l}{CD1}  & \multicolumn{1}{l}{CD2}  & \multicolumn{1}{l|}{NC}    & \multicolumn{1}{l}{CD1}  & \multicolumn{1}{l}{CD2}  & \multicolumn{1}{l|}{NC}    & \multicolumn{1}{l}{CD1} & \multicolumn{1}{l}{CD2} & NC    &  \\ \cline{1-19}
SPSR \cite{kazhdan2013screened}                                  
& 0.178                     & 0.2050                     & \multicolumn{1}{l|}{0.874}
& 0.225                     & 0.2860                     & \multicolumn{1}{l|}{\cellcolor{tabsecond}0.861} 
& 0.280                     & 0.3650                     & \multicolumn{1}{l|}{\cellcolor{tabthird}0.869} 
& 0.300                     & 0.4800                     & \multicolumn{1}{l|}{0.866} 
& 0.588                     & 1.6730                     & \multicolumn{1}{l|}{0.879} 
& 0.314                    & 0.6024                    & 0.870 &  \\
NDrop \cite{boulch2021needrop}                                  
& 0.200                     & 0.1140                     & \multicolumn{1}{l|}{0.825} 
& 0.168                     & 0.0630                     & \multicolumn{1}{l|}{0.696} 
& 0.156                     & 0.0500                     & \multicolumn{1}{l|}{0.663} 
& 0.150                     & 0.0810                     & \multicolumn{1}{l|}{0.815} 
& 0.203                     & 0.1390                     & \multicolumn{1}{l|}{0.844} 
& 0.175                    & 0.0894                   & 0.769 &  \\
NP \cite{ma2020neural}                                   
& 0.064                     & 0.0080                     & \multicolumn{1}{l|}{\cellcolor{tabsecond}0.898} 
& 0.049                     & 0.0050                     & \multicolumn{1}{l|}{0.828} 
& 0.133                     & 0.0380                     & \multicolumn{1}{l|}{0.847} 
& 0.060                     & 0.0050                     & \multicolumn{1}{l|}{0.910} 
& 0.178                     & 0.0240                     & \multicolumn{1}{l|}{0.908}
& 0.097                    & 0.0160                    & 0.878 &  \\
SAP \citep{peng2021shape}                                   
& 0.153                     & 0.1010                     & \multicolumn{1}{l|}{0.807} 
& 0.053                     & 0.0090                     & \multicolumn{1}{l|}{0.771} 
& 0.134                     & 0.0330                     & \multicolumn{1}{l|}{0.813}
& 0.070                     & 0.0070                     & \multicolumn{1}{l|}{0.867}
& 0.474                     & 0.3820                     & \multicolumn{1}{l|}{0.725}
& 0.151                    & 0.1064                    & 0.797 &  \\
NSpline \cite{williams2021neural}                             
& 0.135                     & 0.1230                     & \multicolumn{1}{l|}{\cellcolor{tabthird}0.891} 
& 0.056                     & 0.0230                     & \multicolumn{1}{l|}{\cellcolor{tabthird}0.855}
& 0.063                     & 0.0390                     & \multicolumn{1}{l|}{0.827}
& 0.124                     & 0.0910                     & \multicolumn{1}{l|}{0.897}
& 0.378                     & 0.7680                     & \multicolumn{1}{l|}{0.892} 
& 0.151                    & 0.2088                    & 0.872  &  \\
NTPS \cite{NeuralTPS}                                 
& 0.055                     & 0.0050                     & \multicolumn{1}{l|}{\cellcolor{tabfirst}0.909}
& 0.045                     & \cellcolor{tabsecond}0.0030                     & \multicolumn{1}{l|}{\cellcolor{tabfirst}0.892}
& 0.129                     & 0.0220                     & \multicolumn{1}{l|}{\cellcolor{tabfirst}0.872} 
& 0.054                     & 0.0040                     & \multicolumn{1}{l|}{\cellcolor{tabfirst}0.939}
& 0.103                     & 0.0170                     & \multicolumn{1}{l|}{\cellcolor{tabsecond}0.935}
& 0.077                    & 0.0102                    & \cellcolor{tabfirst}0.897 &  \\ 
NAP \cite{nap}                                   & \multicolumn{1}{r}{0.051} & \multicolumn{1}{r}{0.006} & \multicolumn{1}{r|}{0.881}  & \multicolumn{1}{r}{\cellcolor{tabthird}0.037} & \multicolumn{1}{r}{0.002} & \multicolumn{1}{r|}{0.833}  & \multicolumn{1}{r}{0.044} & \multicolumn{1}{r}{0.011} & \multicolumn{1}{r|}{0.862}  & \multicolumn{1}{r}{0.035} & \multicolumn{1}{r}{\cellcolor{tabthird}0.003} & \multicolumn{1}{r|}{0.912}  & \multicolumn{1}{r}{0.042} & \multicolumn{1}{r}{0.002} & \multicolumn{1}{r|}{0.925} & 0.041                    & 0.004                    & \cellcolor{tabthird}0.881 &  \\
SparseOcc \cite{sparseocc}                                   & \multicolumn{1}{r}{\cellcolor{tabthird}0.022} & \multicolumn{1}{r}{\cellcolor{tabthird}0.001} & \multicolumn{1}{l|}{0.871}  & \multicolumn{1}{r}{0.041} & \multicolumn{1}{r}{0.012} & \multicolumn{1}{l|}{0.812}  & \multicolumn{1}{r}{\cellcolor{tabfirst}0.021} & \multicolumn{1}{r}{\cellcolor{tabfirst}0.001} & \multicolumn{1}{l|}{\cellcolor{tabsecond}0.870}  & \multicolumn{1}{r}{\cellcolor{tabthird}0.028} & \multicolumn{1}{r}{\cellcolor{tabthird}0.003} & \multicolumn{1}{l|}{\cellcolor{tabthird}0.931}  & \multicolumn{1}{r}{\cellcolor{tabthird}0.026} & \multicolumn{1}{r}{\cellcolor{tabthird}0.001} & \multicolumn{1}{l|}{\cellcolor{tabfirst}0.936} & \cellcolor{tabthird}0.027                    & \cellcolor{tabthird}0.003                    & \cellcolor{tabsecond}0.886 &  \\

Ours (WDRO)                                  
& \multicolumn{1}{r}{\cellcolor{tabfirst}0.014}    & \multicolumn{1}{r}{\cellcolor{tabfirst}0.0006} & \multicolumn{1}{r|}{0.871}  
& \multicolumn{1}{r}{\cellcolor{tabsecond}0.028} & \multicolumn{1}{r}{\cellcolor{tabthird}0.0036} & \multicolumn{1}{r|}{0.820}  
& \multicolumn{1}{r}{\cellcolor{tabthird}0.038 }  & \multicolumn{1}{r}{\cellcolor{tabthird}0.0051} & \multicolumn{1}{r|}{0.803}  
& \multicolumn{1}{r}{\cellcolor{tabfirst}0.019} & \multicolumn{1}{r}{\cellcolor{tabfirst}0.0005} & \multicolumn{1}{r|}{0.930}  
& \multicolumn{1}{r}{\cellcolor{tabfirst}0.009} & \multicolumn{1}{r}{\cellcolor{tabfirst}0.0003} & \multicolumn{1}{r|}{\cellcolor{tabfirst}0.936} 
& \cellcolor{tabsecond}0.022                   & \cellcolor{tabsecond}0.0020                   & 0.872 &  \\ 
\textbf{Ours} (SDRO)   
& \multicolumn{1}{r}{\cellcolor{tabsecond}0.015}    & \multicolumn{1}{r}{\cellcolor{tabfirst}0.0006} & \multicolumn{1}{r|}{0.873}  
& \multicolumn{1}{r}{\cellcolor{tabfirst}0.021} & \multicolumn{1}{r}{\cellcolor{tabfirst}0.0017} & \multicolumn{1}{r|}{0.823}  
& \multicolumn{1}{r}{\cellcolor{tabsecond}0.027 }  & \multicolumn{1}{r}{\cellcolor{tabsecond}0.0032} & \multicolumn{1}{r|}{0.842}  
& \multicolumn{1}{r}{\cellcolor{tabsecond}0.021} & \multicolumn{1}{r}{\cellcolor{tabsecond}0.0006} & \multicolumn{1}{r|}{\cellcolor{tabsecond}0.932}  
& \multicolumn{1}{r}{\cellcolor{tabsecond}0.020} & \multicolumn{1}{r}{\cellcolor{tabsecond}0.0005} & \multicolumn{1}{r|}{\cellcolor{tabthird}0.934} 
& \cellcolor{tabfirst}0.020                   & \cellcolor{tabfirst}0.0013                    &\cellcolor{tabthird}0.881 &   \\ \cline{1-19}
\end{tabular}
}
\caption{3D Scene \citep{zhou2013dense} reconstructions from sparse  point clouds.}
\label{tab:3ds}
\end{table*}

\vspace{-5pt}
\subsection{Real articulated shape reconstruction}
\vspace{-5pt}
We evaluate the reconstruction of human shapes from the Faust dataset~\citep{Bogo:CVPR:2014} using sparse, noisy point clouds. Quantitative and qualitative comparisons with competing methods are provided in ~\cref{tab:fs} and ~\cref{fig:fs}, respectively. Across all metrics, our distributionally robust training procedures demonstrate superior performance, with SDRO achieving slightly better accuracy and faster convergence than WDRO. Visually, our reconstructions exhibit a significant improvement, especially in capturing finer details at body extremities, which pose challenges due to sparse input data and can lead to ambiguous shape predictions, similar to the fine structures observed in ShapeNet experiments. Notably, NAP outperforms our approach in this setting, and our method is comparable to SparseOcc, as these methods tend to perform better under low noise conditions, whereas ours is specifically designed for robustness under higher noise levels. In contrast, NTPS reconstructions tend to be coarser with fewer details. It is also worth mentioning that several generalizable methods, particularly those trained on ShapeNet (seen in the upper section of the table), show limited effectiveness in this experiment.

\begin{figure}[t!]
\centering
\includegraphics[width=1.0\linewidth]{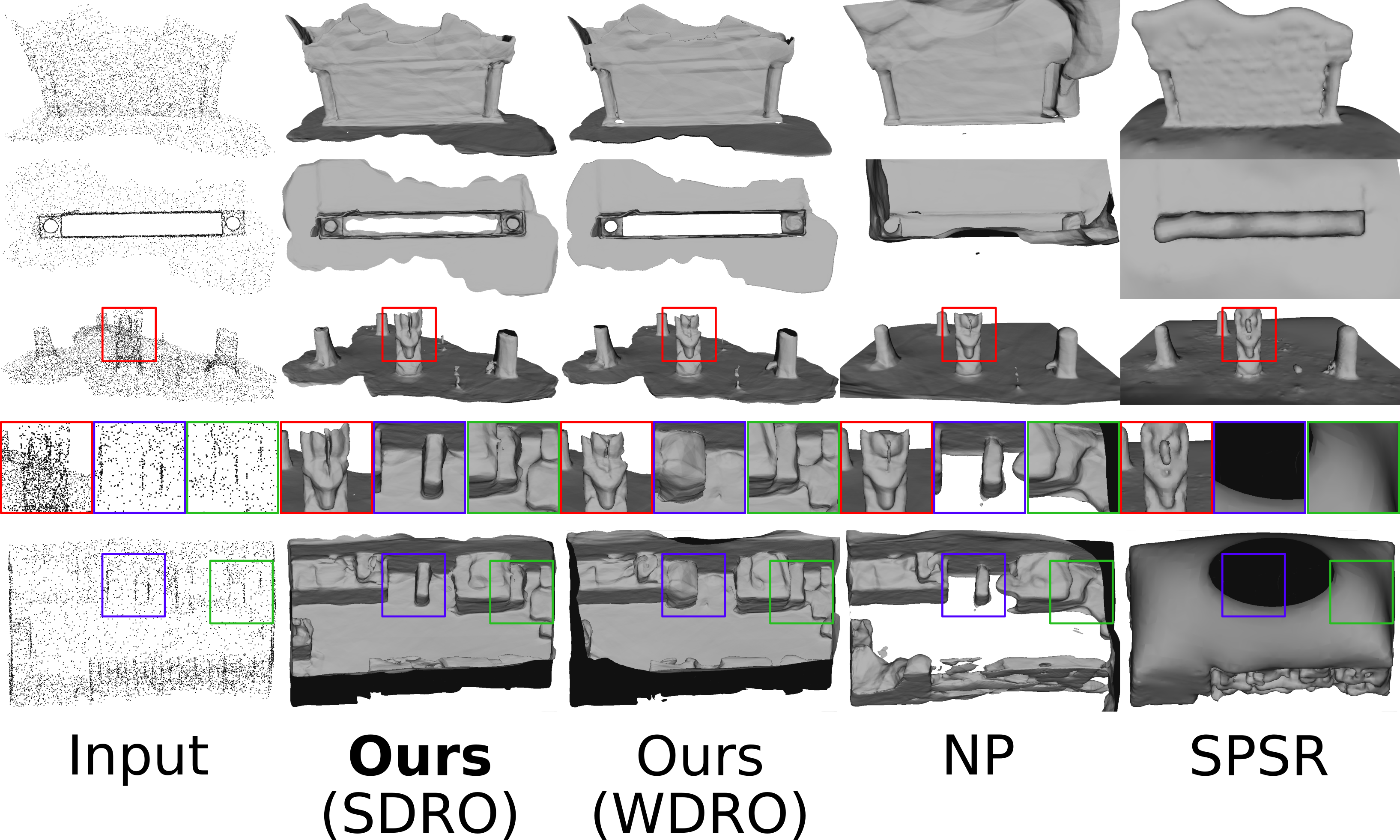}
\vspace{-15pt}
\caption{3D Scene \citep{zhou2013dense} reconstructions from sparse unoriented point clouds.}
\label{fig:3ds}
\end{figure}

\begin{figure}[h!]
\includegraphics[width=1.0\linewidth]{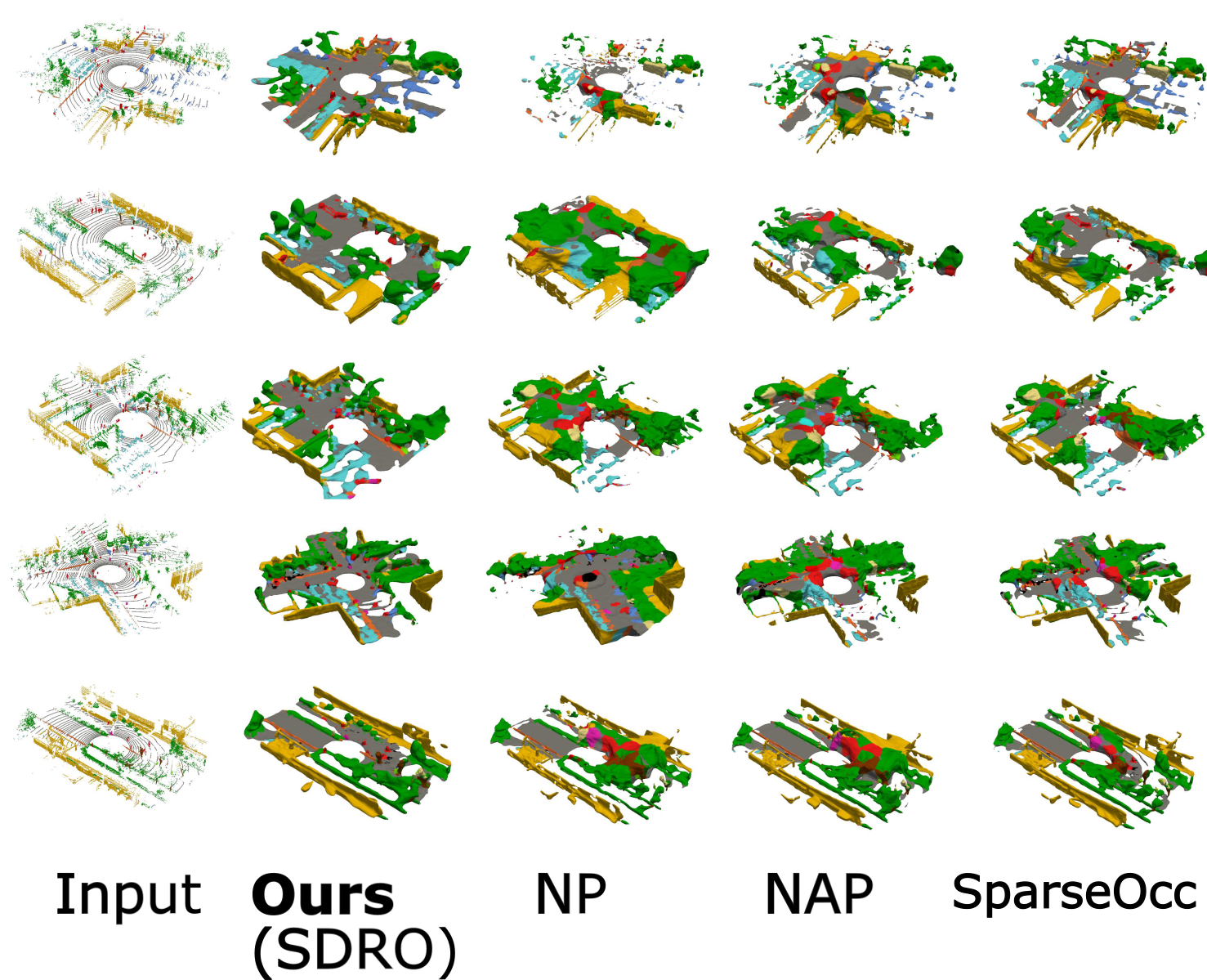}
\vspace{-5pt}
\caption{SemanticPOSS \cite{semantic} reconstruction from road scene LiDAR data.}
\label{fig:lidar}
\end{figure}

\begin{figure}[t!]
\vspace{15pt}
\centering

\includegraphics[width=1.0\linewidth]{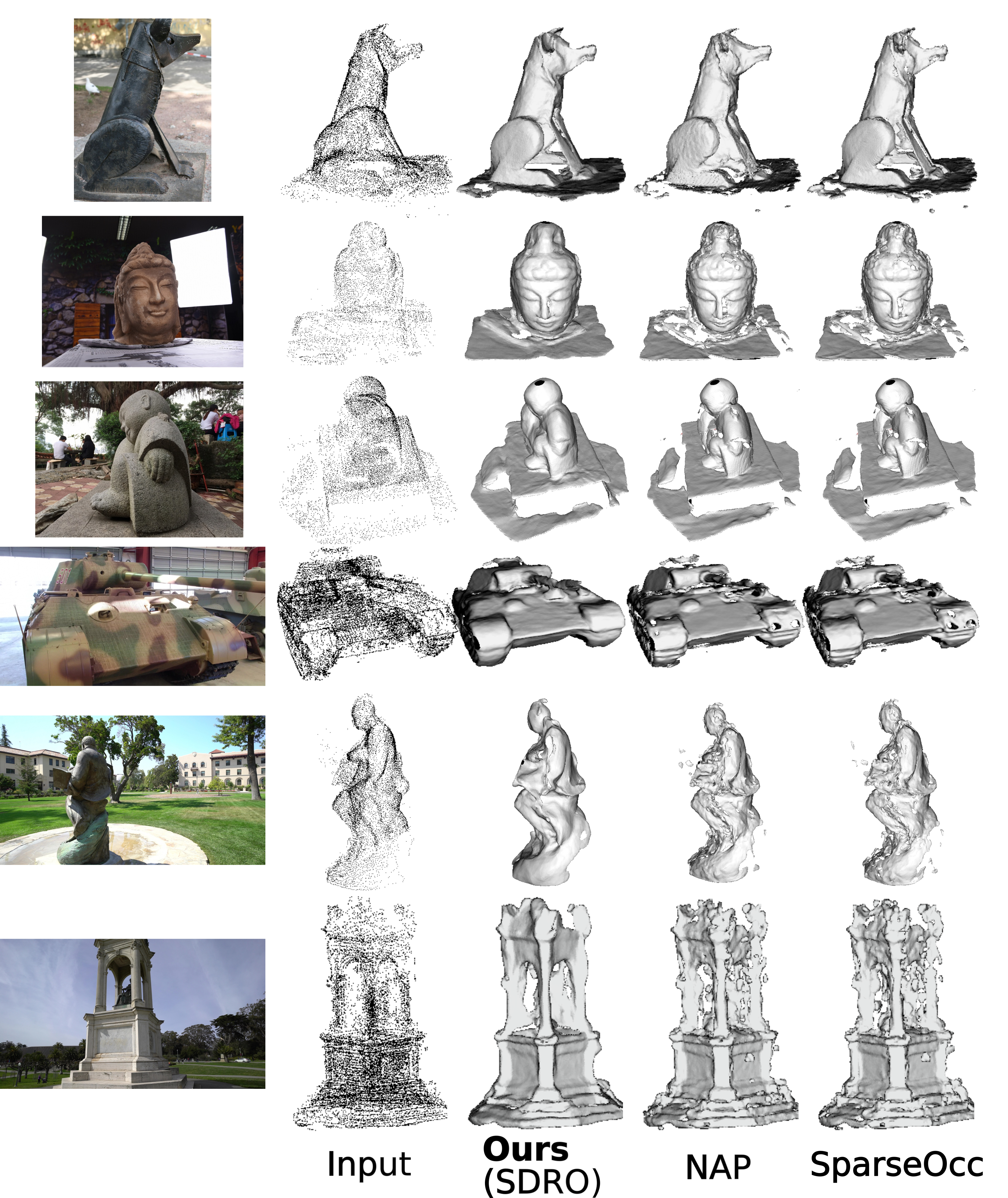}

\caption{Reconstructions from VGGSfM point clouds of sparse views from BlendedMVS \citep{bmvs} and Tanks \& Temples datasets \citep{tanks}. 
}
\label{fig:bmvs_tnt}
\end{figure}
\vspace{-5pt}
\subsection{Real scene level reconstruction}
\vspace{-5pt}
We report reconstruction results on the 3D Scene dataset~\citep{zhou2013dense} from sparse point clouds, following in~\cite{NeuralTPS}. Comparative results for state-of-the-art methods, including NTPS, NP, SAP, NDrop, and NSpline, are obtained from NTPS, while the performances of NAP and SparseOcc are cited from their respective publications and summarized in ~\cref{tab:3ds}. Our method demonstrates superior performance in this setting, attributable to our loss function’s capacity to handle high levels of noise, unlike NAP. Qualitative comparisons with our NP baseline and SPSR are shown in ~\cref{fig:3ds}, where specific regions highlighted by colored boxes illustrate areas where our approach achieves notably high detail and reconstruction fidelity.

Additionally, we conduct qualitative comparisons on BlendedMVS \citep{bmvs} and large-scale scenes from the Tanks \& Temples dataset \citep{tanks} using sparse views. VGGSfM \citep{wang2024vggsfm}, a recent state-of-the-art fully differentiable structure-from-motion pipeline, is used to generate the sparse point cloud inputs for this experiment. Although VGGSfM effectively generates point clouds by triangulating 2D point trajectories and learned camera poses, the sparse input views result in sparse and noisy point clouds, making SDF-based reconstruction challenging. To illustrate the strength of our method, we compare $3$ examples from each dataset against SparseOcc and NAP in \cref{fig:bmvs_tnt}, demonstrating sharper details, especially on large-scale scenes from Tanks \& Temples, where other methods struggle due to noise in VGGSfM’s point clouds.

To further evaluate the robustness of our method, we present reconstruction results on the SemanticPOSS dataset~\cite{semantic} and provide qualitative comparisons with SparseOcc, NAP, and NP in ~\cref{fig:lidar}. The visualizations use the dataset’s color-coded semantic segmentations, which were not utilized during training. Our approach achieves marked improvements in reconstruction quality, largely due to our DRO framework. In particular, elements such as cars, trees, and pedestrians are reconstructed with significantly greater detail and precision, whereas baseline methods often blend these object categories into indistinct forms. Additionally, our SDRO method is notably effective in preserving the broader scene structure. Although SparseOcc and NAP demonstrate solid performance under low-noise conditions, their accuracy deteriorates sharply under higher noise levels. Additional qualitative examples are provided in the supplementary materials.

    


\vspace{-5pt}
\section{Ablation studies}
\vspace{-5pt}

\begin{figure}[t!]
\centering
\begin{minipage}{0.3\textwidth}
\centering
\includegraphics[width=\linewidth]{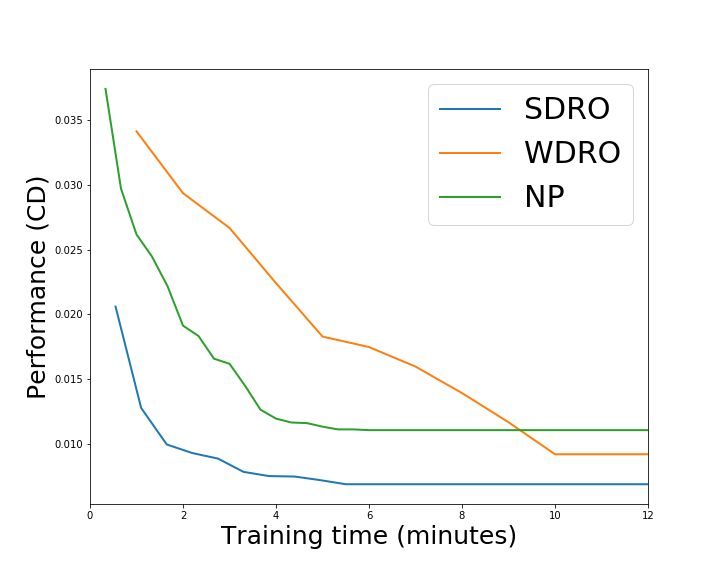}
\vspace{-10pt}
\caption{Performance over training time on Shapenet \citep{shapenet} class Tables.}
\label{fig:abl_time}
\end{minipage}\hspace{5pt}
\begin{minipage}{0.49\textwidth}
\centering
\begin{table}[H]
    \centering
    \scalebox{0.7}{
    \begin{tabular}{c|cc|cc|cc}
    \hline
         & \multicolumn{2}{c}{$\sigma = 0.0$}& \multicolumn{2}{|c}{$\sigma = 0.005$}& \multicolumn{2}{|c}{$\sigma = 0.025$}\\
         &  \textbf{CD1} &  \textbf{NC}&  \textbf{CD1}&  \textbf{NC}&  \textbf{CD1}& \textbf{NC}\\
        \hline
         NP (baseline) \cite{ma2020neural}&  0.73&  0.906&  1.07&  0.847&  2.45& 0.668\\
         NAP \cite{nap} &  \cellcolor{tabthird}0.63&  \cellcolor{tabthird}0.926& \cellcolor{tabthird}0.75&  \cellcolor{tabthird}0.86&  \cellcolor{tabthird}2.21& \cellcolor{tabthird}0.67\\
         SparseOcc \cite{sparseocc} &  \cellcolor{tabsecond}0.56&  \cellcolor{tabsecond}0.931& \cellcolor{tabsecond}0.77&  \cellcolor{tabsecond}0.89&  \cellcolor{tabsecond}2.16& \cellcolor{tabsecond}0.68\\
         Ours(SDRO) & \cellcolor{tabfirst} 0.43& \cellcolor{tabfirst} 0.945&  \cellcolor{tabfirst}0.65&  \cellcolor{tabfirst}0.91& \cellcolor{tabfirst} 1.54& \cellcolor{tabfirst}0.702\\
        \hline
    \end{tabular} }
    \caption{Ablation of our method under varying levels of noise on Shapenet \citep{shapenet} class Tables. }
    \label{tab:noise_ablation}
    
\end{table}
\end{minipage}
\vspace{-15pt}
\end{figure}







\noindent\textbf{Noise ablation} 
To examine the influence of input noise (displacement from the surface) and sparsity on our method’s performance in comparison to the NP baseline, we conduct an ablation study across different noise levels, as shown in ~\cref{tab:noise_ablation}. The results consistently indicate that our method outperforms the baseline at various noise levels. This suggests that our distributionally robust training approach is effective in reducing noise due to both sparse inputs and displacement. Additionally, under high-noise conditions, our method demonstrates superior performance over both NAP and SparseOcc.



\noindent\textbf{Training time} To evaluate the computational efficiency of our approach, we present a performance analysis over training time in ~\cref{fig:abl_time}, comparing our DRO methods with the NP baseline. The plot demonstrates the performance attained after specific training durations. Notably, WDRO achieves baseline performance after approximately $3$ minutes and reaches its peak in $10$ minutes. On the other hand, SDRO shows an advantage over the NP baseline after just $2$ minutes of training, attaining optimal performance in under $6$ minutes, matching the baseline's convergence time while outperforming both the baseline and WDRO. This observation underscores the computational advantages of using the Sinkhorn distance in our distributionally robust optimization formulation (\cref{equ:dro}), as opposed to the Wasserstein distance. Further ablation results can be found in the supplementary material.

\vspace{-5pt}
\section{Limitations}
\label{sec:lim}
\vspace{-5pt}
Our method is able to significantly  improve over recent state of the art methods such as NAP and SparseOcc in very challenging scenarios. However, when the input points are clean and dense, NAP can show competitive or even better performance as shown in \cref{tab:fs} and in the density ablation in the supplementary material. An interesting direction for future research could be to explore strategies that combine local adaptive radii control in NAP with the global control on the worst-case distribution in SDRO. We aim to investigate this as part of our future work. 
\vspace{-5pt}
\section{Conclusion}
\vspace{-5pt}
We have shown that regularizing implicit shape representation learning from sparse unoriented point clouds through distributionally robust optimization with wasserstein uncertainty sets can lead to superior reconstructions. We believe these new findings can usher in a new body of work incorporating distributional robustness in learning neural implicit functions, which in turn can potentially have a larger impact beyond the specific scope of this paper. 



{
    \small
    \bibliographystyle{iee_fullname}
    \bibliography{main}
}





\section{Background on Distributionally Robust Optimization}

Distributionally Robust Optimization (DRO) was initially introduced by \cite{scarf1957min} and has since become a significant framework for addressing uncertainty in decision-making \citep{goh2010distributionally,bertsimas2018data} . The DRO framework operates by defining an uncertainty set $\mathcal{U}$, typically modeled as a ball of radius $\epsilon$ around an empirical distribution $\hat{Q}_n$ , such that $\mathcal{U} = \{Q : d(Q, \hat{Q}_n) \leq \epsilon\}$. The specific choice of the divergence measure  greatly influences both the required size of $\epsilon$ and the tractability of the resulting optimization problem. The loss function is miminized under the worst-case distribution $ Q \in \mathcal{U} $  in  terms of the expected loss.

In machine learning, two primary divergence measures are widely adopted: $f$-divergences and the Wasserstein distance. With  $f$-divergences \citep{ben2013robust, miyato2015distributional, namkoong2016stochastic}  convex optimization techniques are usually leveraged to define tractable uncertainty sets. Alternatively, the Wasserstein distance \citep{wdro,kuhn_dro} is based on a  metric  over the data space, enabling the inclusion of distributions with supports different from the empirical distribution, thereby offering robustness to unseen data. However, the computational complexity of Wasserstein-based DRO makes it more challenging to handle. To address these challenges, various studies have proposed tractable methods for specific uncertainty sets and loss functions. For instance, \citep{wdro,kuhn_dro,shafieezadeh2015distributionally}  provide practical approaches for solving DRO problems with uncertainty regions defined by Wasserstein balls. For smooth loss functions, \cite{wrm} proposes an efficient formulation for certifying robustness under Wasserstein uncertainty sets. Furthermore, the Unified DRO framework (UDR) introduced by \cite{udr} establishes a connection between Wasserstein DRO and adversarial training (AT) methods, offering a novel approach where the dual variable of the DRO problem is adaptively learned during training. This contrasts with  \cite{wrm} , where this parameter is fixed. \citep{sdro, sdro_general} study the DRO problem using the Sinkhorn Distance instead of the Wasserstein Distance providing efficient dual formulations. 

\section{Additional Ablative Analysis}


\begin{table}[h]
\centering
\scalebox{0.9}{
\begin{tabular}{lcc}
\hline
& \textbf{Sparse} & \textbf{Dense} \\
\hline
SPSR \cite{kazhdan2013screened} & 2.27 & 1.25 \\
DIGS \citep{ben2022digs}  & 0.68 & \cellcolor{tabfirst}0.19\\
OG-INR \citep{koneputugodage2023octree}  & 0.85 &  \cellcolor{tabsecond}0.20\\
NTPS \citep{NeuralTPS}  & 0.73 & - \\
NP \citep{ma2020neural}  & 0.58 & 0.23  \\
SparseOcc \citep{sparseocc}  & \cellcolor{tabsecond}0.49 & \cellcolor{tabsecond}0.20  \\
NAP \citep{nap}  & \cellcolor{tabsecond}0.49 & \cellcolor{tabfirst}0.19  \\
Ours (WDRO) & \cellcolor{tabthird}0.51 & \cellcolor{tabsecond}0.20 \\ 
Ours (SDRO) & \cellcolor{tabfirst}0.48 & \cellcolor{tabthird} 0.21 \\ \hline
\end{tabular}}
\caption{\footnotesize Ablation of point cloud density}
\label{tab:srb}
\end{table}


\subsection{Varying the point cloud density}
\begin{figure}[t!]
\centering
\includegraphics[width=1.0\linewidth]{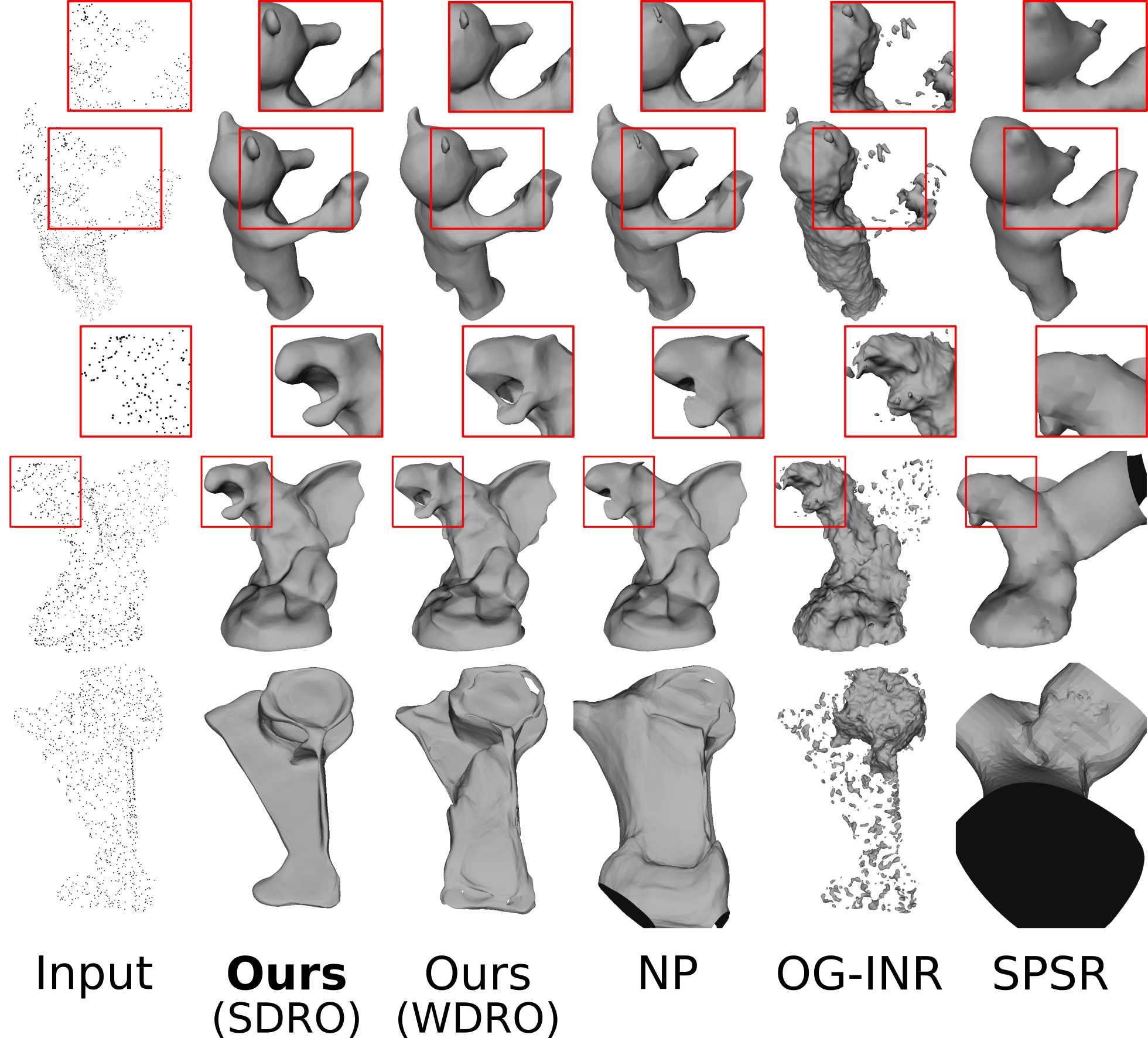}
\vspace{-5pt}
\caption{SRB \citep{williams2019deep} unsupervised reconstructions from sparse (1024 pts) unoriented point clouds without data priors.}
\label{fig:srb}
\vspace{-10pt}
\end{figure}

In order to assess the performance of our method under various point cloud densities we perform an ablative analysis on the SRB benchmark \citep{williams2019deep}. We present quantitative results for both  $1024$-sized and dense input point clouds. In the dense setting, we report results from OG-INR. Our distributionally robust training strategy outperforms competitors in the sparse case and performs on par with the state-of-the-art in the dense case. Importantly, we observe considerable improvement over our baseline (NP) in both scenarios. \cref{fig:srb} visually supports these results, illustrating reconstructions for sparse and dense inputs. In the dense setting, our method captures finer details, emphasized by the red boxes. These results highlight the practical advantages of our approach, even for dense inputs. Interestingly, our ablative analysis reveals that for dense inputs, WDRO may exhibit slightly better performance compared to SDRO. This result is not surprising, given that WDRO is certified  to effectively hedge against small perturbations \citep{wrm}. Consequently, as the input becomes denser, the noise on the labels due to input sparsity diminishes, thereby favoring WDRO.

\subsection{Hyperparameter Analysis}
In order to determine the hyperparameters of our proposed approach (SDRO), We performed a hyperparameter search on the SRB \citep{williams2019deep} benchmark  utilizing the chamfer distance between the reconstruction and the input point cloud as a validation metric. For the remaining datasets, we employed the same hyperparameters. 

We carry out here an ablation study where we vary each one of the hyperparameters $\lambda$ and $\rho$ while fixing the remaining ones in order to better understand the behavior of our approach (SDRO) and its sensitivity to the choice of these hyperparameters.\\

\begin{figure}[t!]
\centering
\begin{minipage}{0.45\textwidth}
\centering
\includegraphics[width=\linewidth]{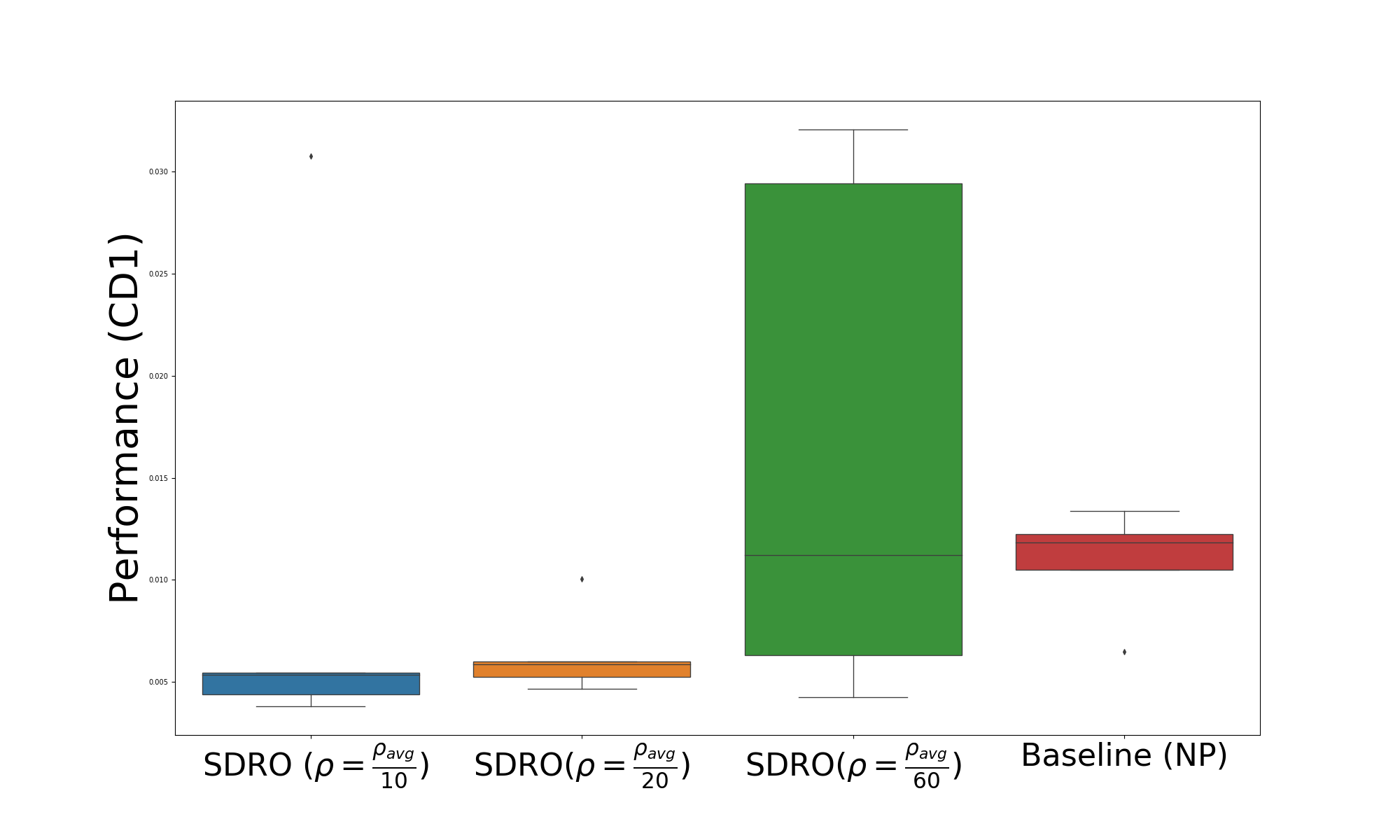}
\vspace{-10pt}
\caption{Ablation of the regularization parameter $\rho$. } 
\label{fig:abl_rho}
\end{minipage}\hspace{5pt}
\begin{minipage}{0.45\textwidth}
\centering
\includegraphics[width=\linewidth]{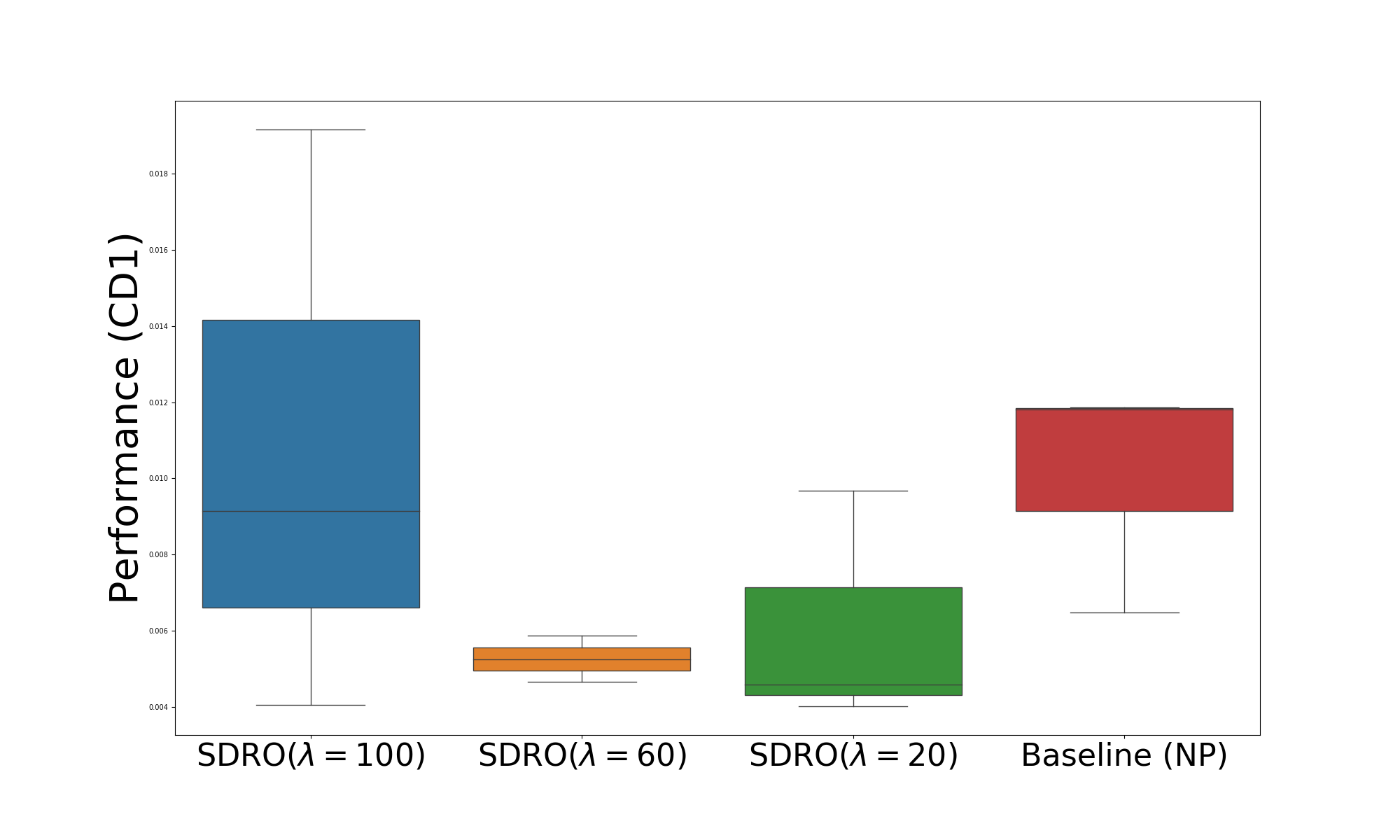}
\caption{Ablation of the regularization parameter $\lambda$. }
\label{fig:abl_lambda}
    
\end{minipage}
\end{figure}

\noindent\textbf{Regularization parameter $\lambda$}. \quad This parameter controls how close the worst-case distribution $Q^\prime$ is to the nominal distribution. \cref{fig:abl_lambda} illustrates how a very high value for this parameter minimizes the regularization impacts of SDRO by maintaining the worst-case samples around the nominal samples. Conversely, excessively low values lead to overly pessimistic estimations over-smoothing the results, despite greatly improving over the NP baseline. \\

\noindent\textbf{Regularization parameter $\rho$}. \quad This parameter is responsible for the strength of the entropic regularization: it controls how the SDRO worst case distribution is concentrated around the support points of WDRO worst case distribution \cite{sdro}. Consequently, it has to be defined such that it facilitates  finding  challenging distributions around the surface while maintaining a useful supervision signal. According to \cref{fig:abl_rho}, it is important to utilize a sufficiently high $\rho$ value in order to hedge against the right family of distributions. Contrastively, very high values can result in increased variance. Notice that $\rho_{avg}$ here corresponds to average  $\sigma_p$ over the input points $\textbf{P}$.

\section{Training algorithm for WDRO}

We provide in \cref{alg:wdro} the detailed training procedure for WDRO.

\begin{algorithm}[t!]
\small
\begin{algorithmic}
\Require Point cloud $\textbf{P}$, learning rate $\alpha$, number of iterations $N_{\text{it}}$, batch size $N_b$.\\
WDRO hyperparameters: $\epsilon$, $\sigma_0$, $\alpha_{wdro}$, $N_{it}^{wdro}$, $\eta_\lambda$. 
\Ensure Optimal parameters ${\theta}^*$.
\State Compute local st. devs. $\{\sigma_p\}$ ($\sigma_p=\max_{t\in K\text{nn}(p,\mathbf{P})}||t-p||_2$).
\State $ \mathfrak{Q} \leftarrow$ sample($\textbf{P}$,$\{\sigma_p\}$). (Equ. \cref{equ:sample})
\State Compute nearest points in $\textbf{P}$ for all samples in  $\mathfrak{Q}$. 
\State Initialize $\lambda_1 = \lambda_2 = 1$.  
\State Initialize $\lambda$.
\For{$N_{\text{it}}$ times}
\State Sample $N_b$ query points   $\{  q, q\sim Q \}$.
\State Initialize $N_b$ points $\{q'\}$, ($q^{\prime} \sim \mathcal{N}(q,\sigma_0\mathbf{I}_3)$).
\For{$N^{wdro}_{\text{it}}$ times}
    \State $q^{\prime} \leftarrow q^{\prime} + \alpha_{wdro} \nabla_{q^{\prime} } [ \mathcal{L}(\theta ,q^{\prime}) - \lambda c(q,q^{\prime} ) ]$
\EndFor
\State $\lambda \leftarrow \lambda-\eta_\lambda\left(\epsilon-\frac{1}{N_b} \sum_{i=1}^{N_b} c\left(q_i^{\prime}, q_i\right)\right)$
\State Compute WDRO losses $\{\mathcal{L}_{\text{WDRO}}(\theta,q)\}$ (Equ. \cref{equ:wdro})
\State Compute combined losses $\{\mathfrak{L}(\theta,q)\}$ (Equ. \cref{equ:final})
\State $(\theta, \lambda_1, \lambda_2) \leftarrow (\theta, \lambda_1, \lambda_2) - \alpha \nabla_{\theta,\lambda_1,\lambda_2} \Sigma_q \mathfrak{L}(\theta ,q)$
\EndFor
\end{algorithmic}
\caption{\small The training procedure of our method with WDRO.}
\label{alg:wdro}
\end{algorithm}

\begin{figure*}[t!]
\centering
\includegraphics[width=0.8 \linewidth]{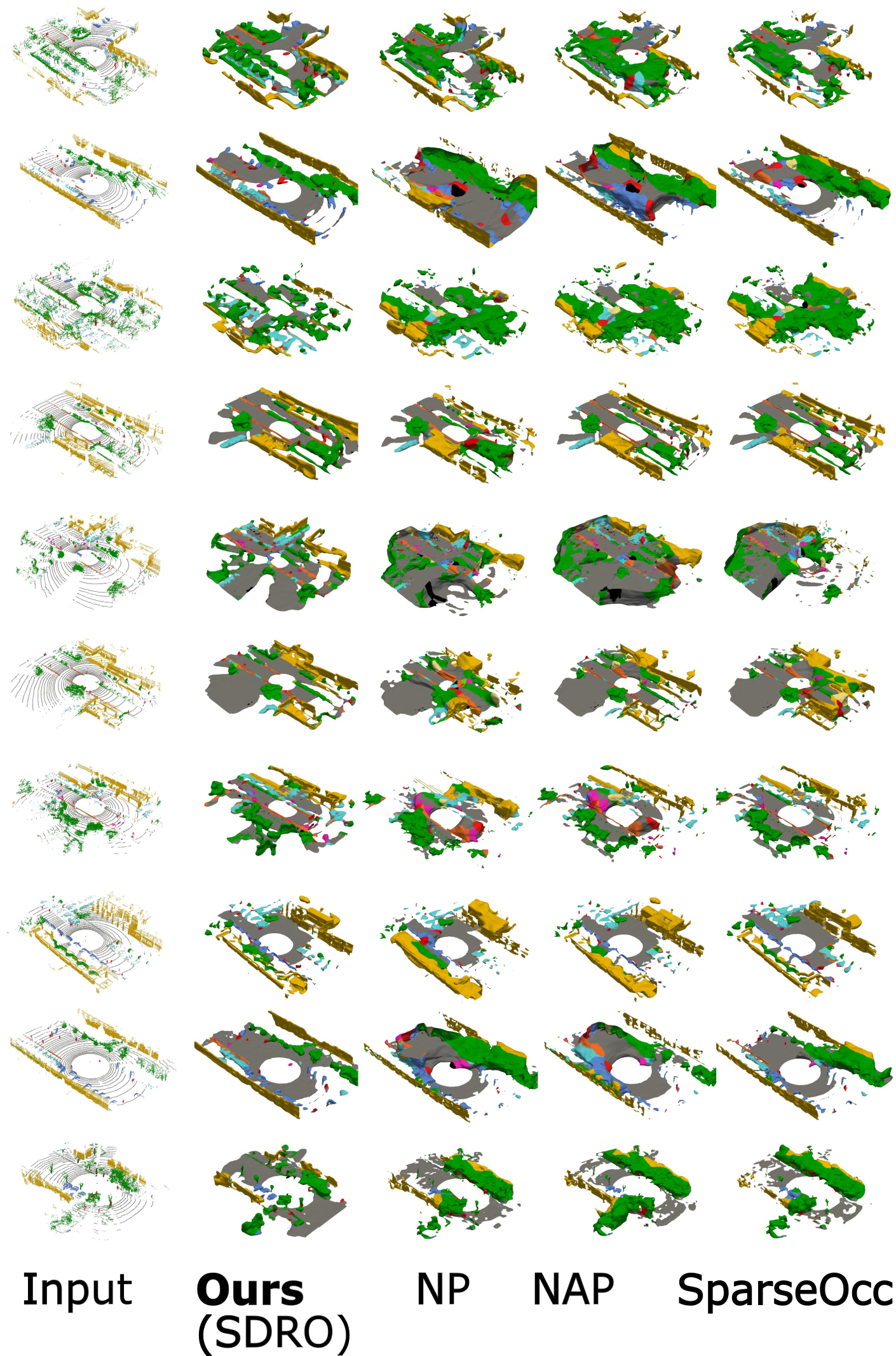}
\vspace{-5pt}
\caption{SemanticPOSS \cite{semantic} reconstructions from road scene LiDAR data.}
\label{fig:lidar_supp}
\vspace{-15pt}
\end{figure*}

\section{Additional Qualitative Results}

We provide additional qualitative comparisons using SemanticPOSS road scene LiDAR data. \cref{fig:lidar_supp} highlights the superiority of our method in this challenging scenario compared to NAP and SparseOcc. This is particularly evident in highly noisy regions, such as trees, where these methods struggle, whereas our SDRO approach demonstrates robust performance.

\section{Evaluation Metrics}


Building on the definitions provided in \cite{boulch2022poco} and \cite{williams2019deep}, we present the formal definitions of the metrics used for evaluation in the main submission. Let $\mathcal{S}$ and  $\hat{\mathcal{S}}$ denote the ground truth and predicted meshes, respectively. Following \cite{NeuralTPS}, all metrics are approximated using 100k samples drawn from $\mathcal{S}$ and $\hat{\mathcal{S}}$  for ShapeNet and Faust, and 1M samples for 3DScene. For SRB, we also utilize 1M samples, as suggested by \cite{ben2022digs} and \cite{koneputugodage2023octree}.

\noindent \textbf{Chamfer Distance (CD1)} The L$_1$  Chamfer Distance is computed using the two-way nearest-neighbor distance:: 
$$\mathrm{CD}_1=\frac{1}{2|\mathcal{S}|} \sum_{v \in \mathcal{S}} \min _{\hat{v} \in \hat{\mathcal{S}}}\|v-\hat{v}\|_2+\frac{1}{2|\hat{\mathcal{S}}|} \sum_{\hat{v} \in \hat{\mathcal{S}}} \min _{v \in \mathcal{S}}\|\hat{v}-v\|_2.$$

\noindent \textbf{Chamfer Distance (CD2)} The L$_2$ Chamfer Distance is computed using the two-way nearest-neighborr squared distance: 
$$\mathrm{CD}_2=\frac{1}{2|\mathcal{S}|} \sum_{v \in \mathcal{S}} \min _{\hat{v} \in \hat{\mathcal{S}}}\|v-\hat{v}\|_2^2+\frac{1}{2|\hat{\mathcal{S}}|} \sum_{\hat{v} \in \hat{\mathcal{S}}} \min _{v \in \mathcal{S}}\|\hat{v}-v\|_2^2.$$

\noindent \textbf{F-Score (FS)} For a given threshold $\tau$, the F-Score between the ground truth mesh  $\mathcal{S}$ and the predicted mesh $\hat{\mathcal{S}}$ is defined as:
$$
\mathrm{FS}\left(\tau, \mathcal{S}, \hat{\mathcal{S}}\right)=\frac{2 \text { Recall} \cdot \text{Precision }}{\text { Recall }+\text { Precision }},
$$

where
$$
\begin{array}{r}
\operatorname{Recall}\left(\tau, \mathcal{S}, \hat{\mathcal{S}}\right)=\mid\left\{v \in \mathcal{S} \text {, s.t. } \min _{\hat{v} \in \hat{ \mathcal{S} }} \left\|v-\hat{v}\|_2\right<\tau\right\} \mid ,\\
\operatorname{Precision}\left(\tau, \mathcal{S}, \hat{\mathcal{S}}\right)=\mid\left\{\hat{v} \in \hat{\mathcal{S} }\text {, s.t. } \min _{v \in  \mathcal{S} } \left\|v-\hat{v}\|_2\right<\tau\right\} \mid .\\
\end{array}
$$
Following \cite{mescheder2019occupancy} and \cite{peng2020convolutional}, we set $\tau$ to $0.01$.

\noindent \textbf{Normal consistency (NC)} measures the alignment of surface normals between two meshes
$\mathcal{S}$ (ground truth)  and  $\hat{\mathcal{S}}$ (prediction). Denoting the normal at a point $v$ in $\mathcal{S}$  by $n_v$, it is defined as

$$\mathrm{NC}=\frac{1}{2|\mathcal{S}|} \sum_{v \in \mathcal{S}} n_{v} \cdot n_{\operatorname{closest}(v,\hat{ \mathcal{S}})}+\frac{1}{2|\hat{\mathcal{S}}|} \sum_{\hat{v} \in \hat{\mathcal{S}}} n_{\hat{v}} \cdot n_{\operatorname{closest}(\hat{v}, \mathcal{S})},$$

where  
$$
\operatorname{closest}(v, \hat{\mathcal{S}}) = \operatorname{argmin} _{\hat{v} \in \hat{\mathcal{S}}}\|v-\hat{v}\|_2.
$$


\end{document}